\documentclass{article}

\usepackage{microtype}
\usepackage{graphicx}
\usepackage{todonotes}
\usepackage{hyperref}
\usepackage{subcaption}

\usepackage{booktabs}
\usepackage{amsfonts,amsmath,amssymb}
\newcommand{\bx}{\mathbf{x}}
\newcommand{\bz}{\mathbf{z}}
\newcommand{\bh}{\mathbf{h}}

\newcommand{\R}{\mathbb{R}}
\usepackage{color}

\usepackage{listings}
\definecolor{mygreen}{rgb}{0,0.5,0}
\definecolor{mygray}{rgb}{0.5,0.5,0.5}
\definecolor{mymauve}{rgb}{0.25,0.5,0.5}

\usepackage[accepted]{icml2018}
\icmltitlerunning{Neural Relational Inference for Interacting Systems}

\begin{document}

\twocolumn[
\icmltitle{Neural Relational Inference for Interacting Systems}
\icmlsetsymbol{equal}{*}

\begin{icmlauthorlist}
\icmlauthor{Thomas Kipf}{equal,am}
\icmlauthor{Ethan Fetaya}{equal,to,vi}
\icmlauthor{Kuan-Chieh Wang}{to,vi}
\icmlauthor{Max Welling}{am,ci}
\icmlauthor{Richard Zemel}{to,vi,ci}
\end{icmlauthorlist}

\icmlaffiliation{to}{University of Toronto, Toronto, Canada}
\icmlaffiliation{am}{University of Amsterdam, Amsterdam, The Netherlands}
\icmlaffiliation{ci}{Canadian Institute for Advanced Research, Toronto, Canada}
\icmlaffiliation{vi}{Vector Institute, Toronto, Canada}

\icmlcorrespondingauthor{Thomas Kipf}{t.n.kipf@uva.nl}

\icmlkeywords{Machine Learning, ICML}

\vskip 0.3in
]

\printAffiliationsAndNotice{\icmlEqualContribution}

\begin{abstract}
Interacting systems are prevalent in nature, from dynamical systems in physics to complex societal dynamics. The interplay of components can give rise to complex behavior, which can often be explained using a simple model of the system's constituent parts. In this work, we introduce the neural relational inference (NRI) model: an unsupervised model that learns to infer interactions while simultaneously learning the dynamics purely from observational data. Our model takes the form of a variational auto-encoder, in which the latent code represents the underlying interaction graph and the reconstruction is based on graph neural networks. In experiments on simulated physical systems, we show that our NRI model can accurately recover ground-truth interactions in an unsupervised manner. We further demonstrate that we can find an interpretable structure and predict complex dynamics in real motion capture and sports tracking data.

\end{abstract}

\section{Introduction}
A wide range of dynamical systems in physics, biology, sports, and other areas can be seen as groups of interacting components, giving rise to complex dynamics at the level of individual constituents and in the system as a whole. Modeling these type of dynamics is challenging: often, we only have access to individual trajectories, without knowledge of the underlying interactions or dynamical model.

As a motivating example, let us take the movement of basketball players on the court. It is clear that the dynamics of a single basketball player are influenced by the other players, and observing these dynamics as a human, we are able to reason about the different types of interactions that might arise, e.g. defending a player or setting a screen for a teammate. It might be feasible, though tedious, to manually annotate certain interactions given a task of interest. It is more promising to learn the underlying interactions, perhaps shared across many tasks, in an unsupervised fashion.

\begin{figure}[tp]
  \centering
  \includegraphics[width=0.8\linewidth]{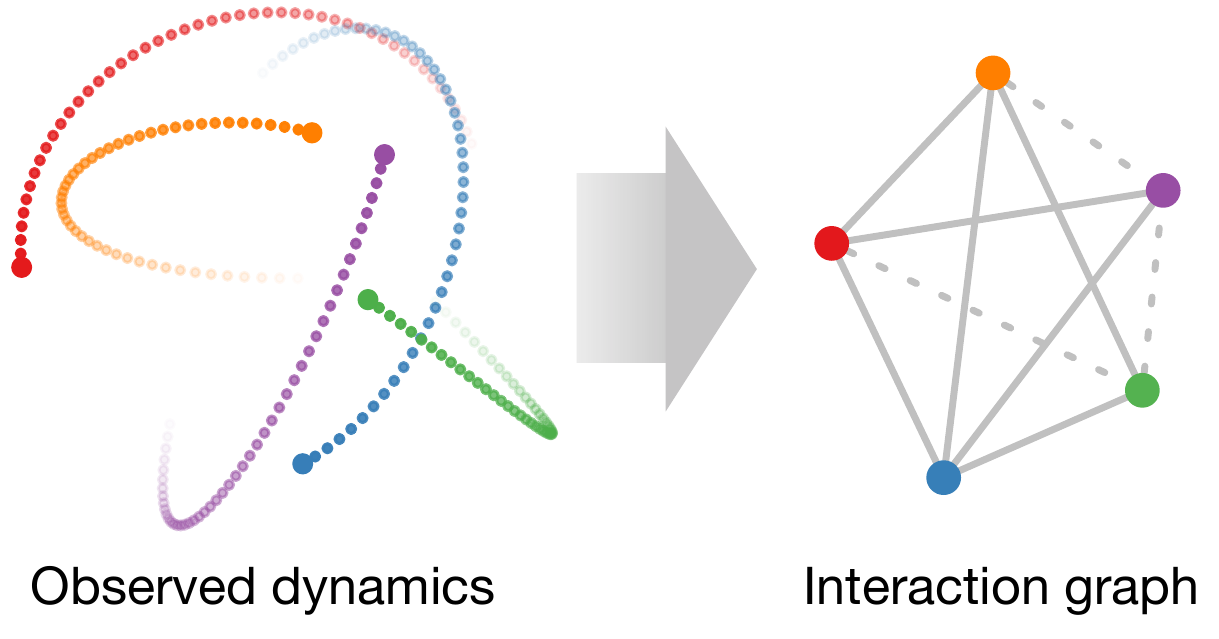}
  \vspace{-1em}
  \caption{Physical simulation of 2D particles coupled by invisible springs (\textit{left}) according to a latent interaction graph (\textit{right}). In this example, solid lines between two particle nodes denote connections via springs whereas dashed lines denote the absence of a coupling. In general, multiple, directed edge types -- each with a different associated relation -- are possible.}
  \label{fig:intro}
\end{figure}

Recently there has been a considerable amount of work on learning the dynamical model of interacting systems using \emph{implicit} interaction models \cite{sukhbaatar2016learning,guttenberg2016permutation,SantoroRBMPBL17,WattersZWBPT17,hoshen2017vain,steenkiste2018relational}. These models can be seen as graph neural networks (GNNs) that send messages over the fully-connected graph, where the interactions are modeled implicitly by the message passing function \cite{sukhbaatar2016learning,guttenberg2016permutation,SantoroRBMPBL17,WattersZWBPT17} or with the help of an attention mechanism \cite{hoshen2017vain,steenkiste2018relational}.

In this work, we address the problem of inferring an \emph{explicit} interaction structure while simultaneously learning the dynamical model of the interacting system in an unsupervised way. Our neural relational inference (NRI) model learns the dynamics with a GNN over a discrete \emph{latent} graph, and we perform inference over these latent variables. The inferred edge types correspond to a clustering of the interactions. Using a probabilistic model allows us to incorporate prior beliefs about the graph structure, such as sparsity, in a principled manner.

In a range of experiments on physical simulations, we show that our NRI model possesses a favorable inductive bias that allows it to discover ground-truth physical interactions with high accuracy in a completely unsupervised way. We further show on real motion capture and NBA basketball data that our model can learn a very small number of edge types that enable it to
accurately predict the dynamics many time steps into the future.

\section{Background: Graph Neural Networks}
We start by giving a brief introduction to a recent class of neural networks that operate directly on graph-structured data by passing local messages \cite{ScarselliGTHM09,YujiaTBZ16,GilmerSRVD17}. We refer to these models as graph neural networks (GNN). Variants of GNNs have been shown to be highly effective at relational reasoning tasks \cite{SantoroRBMPBL17}, modeling interacting or multi-agent systems \cite{sukhbaatar2016learning,BattagliaPLRK16},  classification of graphs \cite{bruna2013spectral,DuvenaudMABHAA15,dai2016discriminative,niepert2016learning,DefferrardBV16,kearnes2016molecular} and classification of nodes in large graphs \cite{KipfW16,hamilton2017inductive}. The expressive power of GNNs has also been studied theoretically in \cite{zaheer2017deep,HerzigRCBG2018}.

\begin{figure}[tp]
  \centering
  \includegraphics[width=\linewidth,trim={0 0.3cm 0.2cm 0},clip]{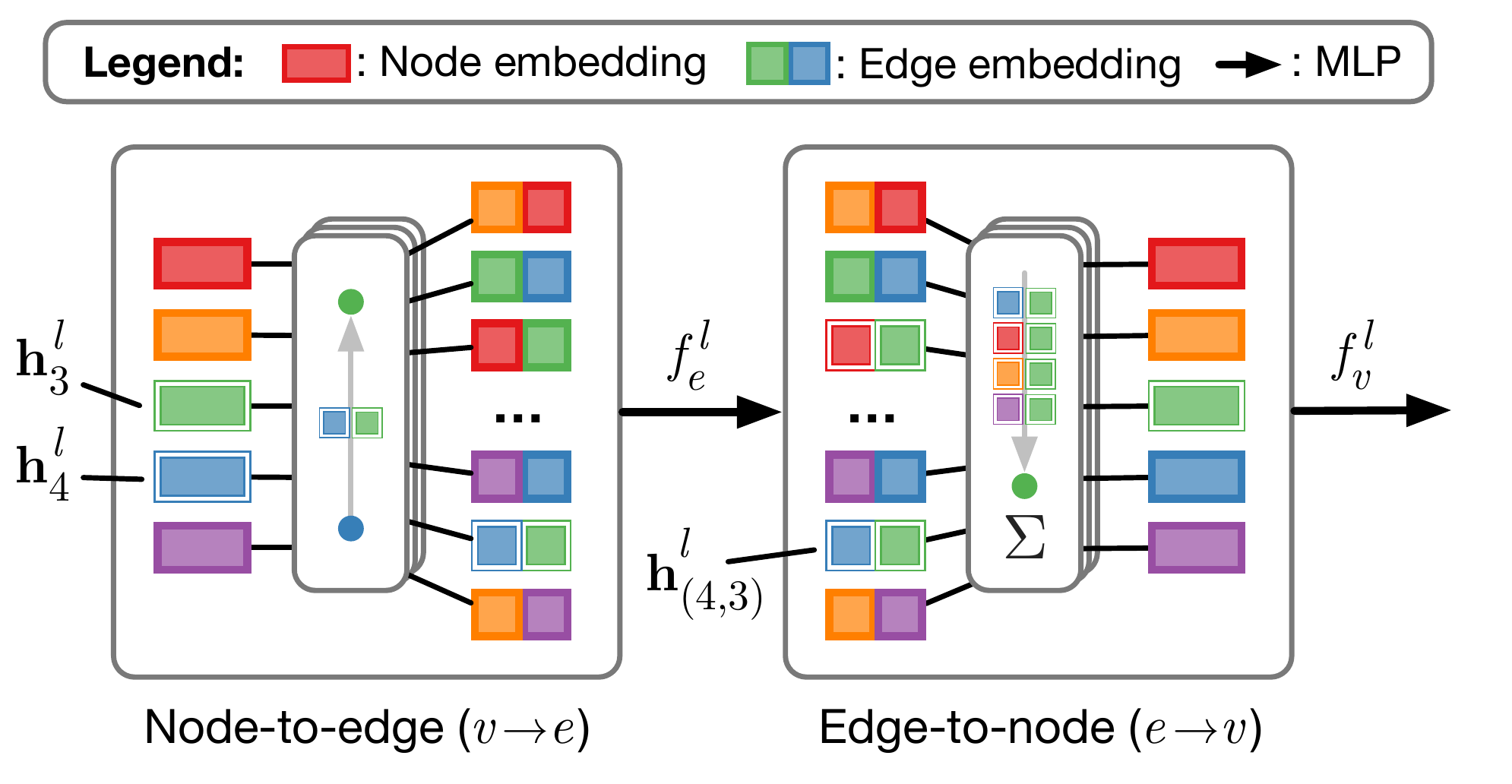}
  \vspace{-1em}
  \caption{Node-to-edge ($v{\rightarrow}e$) and edge-to-node ($e{\rightarrow}v$) operations for moving between node and edge representations in a GNN. $v{\rightarrow}e$ represents concatenation of node embeddings connected by an edge, whereas $e{\rightarrow}v$ denotes the aggregation of edge embeddings from all incoming edges. In our notation in Eqs.~\eqref{eq:gnn_simple1}--\eqref{eq:gnn_simple2}, every such operation is followed by a small neural network (e.g.~a 2-layer MLP), here denoted by a black arrow. For clarity, we highlight which node embeddings are combined to form a specific edge embedding ($v{\rightarrow}e$) and which edge embeddings are aggregated to a specific node embedding ($e{\rightarrow}v$).}
\label{fig:n2e_e2n}
\end{figure}

Given a graph $\mathcal{G}=(\mathcal{V},\mathcal{E})$ with vertices $v\in\mathcal{V}$ and edges $e=(v,v')\in\mathcal{E}$ \footnote{Undirected graphs can be modeled by explicitly assigning two directed edges in opposite direction for each undirected edge.}, we define a single node-to-node message passing operation in a GNN as follows, similar to \citet{GilmerSRVD17}:
\begin{align}
v{\rightarrow}e:\,\,\, \bh^l_{(i,j)} &= f_{e}^l([\bh^l_i,\bh^l_j,\bx_{(i,j)}]) \label{eq:gnn_simple1} \\
e{\rightarrow}v:\,\,\,\, \bh^{l+1}_{j} &= f_{v}^l([\textstyle\sum_{i\in \mathcal{N}_j} \bh^l_{(i,j)},\bx_{j}]) \label{eq:gnn_simple2}
\end{align}
where $\bh^l_i$ is the embedding of node $v_i$ in layer $l$, $\bh^l_{(i,j)}$ is an embedding of the edge $e_{(i,j)}$, and $\bx_i$ and $\bx_{(i,j)}$ summarize initial (or auxiliary) node and edge features, respectively (e.g.~node input and edge type). $\mathcal{N}_j$ denotes the set of indices of neighbor nodes connected by an incoming edge and $[\cdot,\cdot]$ denotes concatenation of vectors. The functions $f_{v}$ and $f_{e}$ are node- and edge-specific neural networks (e.g.~small MLPs) respectively (see Figure \ref{fig:n2e_e2n}). Eqs.~\eqref{eq:gnn_simple1}--\eqref{eq:gnn_simple2} allow for the composition of models that map from edge to node representations or vice-versa via multiple rounds of message passing.

In the original GNN formulation from \citet{ScarselliGTHM09} the node embedding $\bh^l_{(i,j)}$ depends only on $\bh^l_{i}$, the embedding of the sending node, and the edge type, but not on  $\bh^l_{j}$, the embedding of the receiving node. This is of course a special case of this formulation, and more recent works such as interaction networks \cite{BattagliaPLRK16} or message passing neural networks \cite{GilmerSRVD17} are in line with our more general formulation. We further note that some recent works factor $f_e^l(\cdot)$ into a product of two separate functions, one of which acts as a gating or attention mechanism \cite{monti2017geometric,duan2017one,hoshen2017vain,velickovic2018graph,garcia2018few,steenkiste2018relational} which in some cases can have computational benefits or introduce favorable inductive biases.

\begin{figure*}[tp]
  \includegraphics[width=\textwidth,trim={0 0.2cm 0 0},clip]{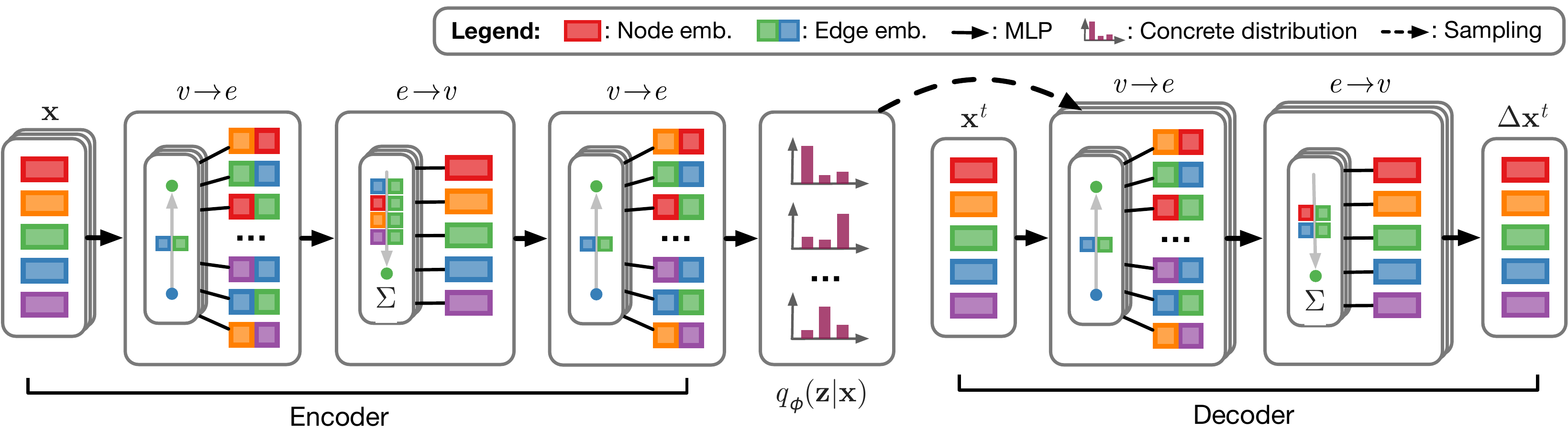}
  \caption{The NRI model consists of two jointly trained parts: An encoder that predicts a probability distribution $q_{\phi}(\bz|\bx)$ over the latent interactions given input trajectories; and a decoder that generates trajectory predictions conditioned on both the latent code of the encoder and the previous time step of the trajectory. The encoder takes the form of a GNN with multiple rounds of node-to-edge ($v{\rightarrow}e$) and edge-to-node ($e{\rightarrow}v$) message passing, whereas the decoder runs multiple GNNs in parallel, one for each edge type supplied by the latent code of the encoder $q_{\phi}(\bz|\bx)$. }
\label{fig:model}
\end{figure*}

\section{Neural Relational Inference Model}\label{sec:GNN}
Our NRI model consists of two parts trained jointly: An encoder that predicts the interactions given the trajectories, and a decoder that learns the dynamical model given the interaction graph.

More formally, our input consists of trajectories of $N$ objects. We denote by $\bx_{i}^t$ the feature vector of object $v_i$ at time $t$, e.g.~location and velocity. We denote by $\bx^t=\{\bx^t_1,...,\bx^t_N\}$ the set of features of all $N$ objects at time $t$, and we denote by $\bx_i=(\bx_i^1,...,\bx_i^T)$ the trajectory of object $i$, where $T$ is the total number of time steps. Lastly, we mark the whole trajectories by $\bx=(\bx^1,...,\bx^T)$. We assume that the dynamics can be modeled by a GNN given an unknown graph $\bz$ where $\bz_{ij}$ represents the discrete edge type between objects $v_i$ and $v_j$. The task is to simultaneously learn to predict the edge types and learn the dynamical model in an unsupervised way.

We formalize our model as a variational autoencoder (VAE) \cite{DiederikW14,rezende2014stochastic} that maximizes the ELBO:
\begin{align}
\mathcal{L}=\mathbb{E}_{q_{\phi}(\bz|\bx)}[\log p_\theta(\bx|\bz)]-\mathrm{KL}[q_\phi(\bz|\bx)||p_\theta(\bz)]
\label{eq:elbo}
\end{align}
The encoder $q_\phi(\bz|\bx)$ returns a factorized distribution of $\bz_{ij}$, where $\bz_{ij}$ is a discrete categorical variable representing the edge type between object $v_i$ and $v_j$. We use a one-hot representation of the $K$ interaction types for  $\bz_{ij}$.

The decoder
\begin{align}
p_\theta(\bx|\bz)=\textstyle\prod_{t=1}^Tp_\theta(\bx^{t+1}|\bx^t,...,\bx^1,\bz)
\end{align}
models $p_\theta(\bx^{t+1}|\bx^t,...,\bx^1,\bz)$ with a GNN given the latent graph structure $\bz$.

The prior $p_\theta(\bz)=\prod_{i\neq j}p_\theta(\bz_{ij})$ is a factorized uniform distribution over edges types. If one edge type is ``hard coded" to represent ``non-edge" (no messages being passed along this edge type), we can use an alternative prior with higher probability on the ``non-edge" label. This will encourage sparser graphs.

There are some notable differences between our model and the original formulation of the VAE \cite{DiederikW14}. First, in order to avoid the common issue in VAEs of the decoder ignoring the latent code $\bz$ \cite{chenKSDDSSA2017}, we train the decoder to predict multiple time steps and not a single step as the VAE formulation requires. This is necessary since interactions often only have a small effect in the time scale of a single time step. Second, the latent distribution is discrete, so we use a continuous relaxation in order to use the reparameterization trick. Lastly, we note that we do not learn the probability $p(\bx^1)$ (i.e.~for $t=1$) as we are interested in the dynamics and interactions, and this does not have any effect on either (but would be easy to include if there was a need).

The overall model is schematically depicted in Figure \ref{fig:model}. In the following, we describe the encoder and decoder components of the model in detail.

\subsection{Encoder}
At a high level, the goal of the encoder is to infer pairwise interaction types $\bz_{ij}$ given observed trajectories $\bx=(\bx^1,...,\bx^T)$.  Since we do not know the underlying graph, we can use a GNN on the fully-connected graph to predict the latent graph structure.

More formally, we model the encoder as $q_{\phi}(\bz_{ij}|\bx)=\mathrm{softmax}(f_{\mathrm{enc}, \phi}(\bx)_{ij,1:K})$, where $f_{\mathrm{enc}, \phi}(\bx)$ is a GNN acting on the fully-connected graph (without self-loops). Given input trajectories $\bx_1,...,\bx_K$ our encoder computes the following message passing operations:
\begin{align}
\bh^1_{j} &= f_{\mathrm{emb}}(\bx_j) \label{eq:enc1} \\
v{\rightarrow}e:\quad\bh^1_{(i,j)} &= f_{e}^1([\bh^1_i,\bh^1_j]) \label{eq:enc2} \\
e{\rightarrow}v:\quad\quad\,\bh^2_{j} &= f_{v}^1(\textstyle\sum_{i\neq j}\bh^1_{(i,j)}) \label{eq:enc3} \\
v{\rightarrow}e:\quad\bh^2_{(i,j)} &= f_{e}^2([\bh^2_i,\bh^2_j])
\label{eq:enc4}
\end{align}
Finally, we model the edge type posterior as $q_{\phi}(\bz_{ij}|\bx) = \mathrm{softmax}(\bh^2_{(i,j)})$ where $\phi$ summarizes the parameters of the neural networks in Eqs.~\eqref{eq:enc1}--\eqref{eq:enc4}. The use of multiple passes, two in the model presented here, allows the model to ``disentangle" multiple interactions while still using only binary terms. In a single pass, Eqs.~\eqref{eq:enc1}--\eqref{eq:enc2}, the embedding $\bh^1_{(i,j)}$ only depends on $\bx_i$ and $\bx_j$  ignoring interactions with other nodes, while  $\bh^2_{j}$ uses information from the whole graph.

The functions $f_{(\ldots)}$ are neural networks that map between the respective representations. In our experiments we used either fully-connected networks (MLPs) or 1D convolutional networks (CNNs) with attentive pooling similar to \cite{lin2017structured} for the  $f_{(\ldots)}$ functions. See supplementary material for further details.

While this model falls into the general framework presented in Sec.~\ref{sec:GNN}, there is a conceptual difference in how $\bh^l_{(i,j)}$ are interpreted. Unlike in a typical GNN, the messages $\bh^l_{(i,j)}$ are no longer considered just a transient part of the computation, but an integral part of the model that represents the edge embedding used to perform edge classification.

\subsection{Sampling}
It is straightforward to sample from $q_{\phi}(\bz_{ij}|\bx)$, however we cannot use the reparametrization trick to backpropagate though the sampling as our latent variables are discrete. A recently popular approach to handle this difficulty is to sample from a continuous approximation of the discrete distribution    \cite{maddisonMT2017,JangGP2017} and use the repramatrization trick to get (biased) gradients from this approximation. We used the concrete distribution \cite{maddisonMT2017} where samples are drawn as:
\begin{align}
\bz_{ij} = \mathrm{softmax}((\bh^2_{(i,j)} + \mathbf{g})/\tau)
\end{align}
where $\mathbf{g}\in\R^K$ is a vector of i.i.d.~samples drawn from a $\mathrm{Gumbel}(0,1)$ distribution and $\tau$ (softmax temperature) is a parameter that controls the ``smoothness" of the samples. This distribution converges to one-hot samples from our categorical distribution when $\tau\rightarrow 0$.

\subsection{Decoder} \label{sec:decoder}
The task of the decoder is to predict the future continuation of the interacting system's dynamics $p_\theta(\bx^{t+1}|\bx^t,...,\bx^1,\bz)$. Since the decoder is conditioned on the graph $\bz$ we can in general use any GNN algorithm as our decoder.

For physics simulations the dynamics is Markovian $p_\theta(\bx^{t+1}|\bx^t,...,\bx^1,\bz)=p_\theta(\bx^{t+1}|\bx^t,\bz)$, if the state is location and velocity and $\bz$ is the ground-truth graph. For this reason we use a GNN similar to interaction networks; unlike interaction networks we have a separate neural network for each edge type. More formally:
\begin{align}
v{\rightarrow}e:\,\,\,\tilde{\bh}^{t}_{(i,j)} &= \sum_k z_{ij,k} \tilde{f}^k_{e}([\bx^t_i,\bx^t_j]) \label{eq:dec_v2e}\\
e{\rightarrow}v:\quad\,\,\boldsymbol{\mu}^{t+1}_{j} &= \bx^t_j+\tilde{f}_{v}(\textstyle\sum_{i\neq j}\tilde{\bh}^{t}_{(i,j)})  \label{eq:dec_e2v}\\
 p(\bx_j^{t+1}|\bx^t,\bz) &= \mathcal{N}(\boldsymbol{\mu}_j^{t+1},\sigma^2 \mathbf{I})
\label{eq:dec3}
\end{align}
Note that $z_{ij,k}$ denotes the $k$-th element of the vector $\bz_{ij}$ and $\sigma^2$ is a fixed variance. When $z_{ij,k}$ is a discrete one-hot sample the messages $\tilde{\bh}^{t}_{(i,j)}$ are $\tilde{f}^k_{e}([\bx^t_i,\bx^t_j])$ for the selected edge type $k$,  and for the continuous relaxation we get a weighted sum. Also note that since in Eq.~\ref{eq:dec_e2v} we add the present state $\bx_j^t$ our model only learns the change in state $\Delta \bx_j^t$.

\subsection{Avoiding degenerate decoders}\label{sec:degenerate}
If we look at the ELBO, Eq.~\ref{eq:elbo}, the reconstruction loss term has the form $\sum_{t=1}^T\log[p(\bx^t|\bx^{t-1},\bz)]$ which involves only single step predictions. One issue with optimizing this objective is that the interactions can have a small effect on short-term dynamics. For example, in physics simulations a fixed velocity assumption can be a good approximation for a short time period. This leads to a sub-optimal decoder that ignores the latent edges completely and achieves only a marginally worse reconstruction loss.

We address this issue in two ways: First, we predict multiple steps into the future, where a ``degenerate" decoder (which ignores the latent edges) would perform much worse. Second, instead of having one neural network that computes the messages given $[\bx^t_i,\bx^t_j,\bz_{ij}]$, as was done in \cite{BattagliaPLRK16}, we have a separate MLP for each edge type. This makes the dependence on the edge type more explicit and harder  to be ignored by the model.

Predicting multiple steps is implemented by replacing the correct input $\bx^t$, with the predicted mean $\boldsymbol{\mu}^t$ for $M$ steps (we used $M=10$ in our experiments), then feed in the correct previous step and reiterate. More formally, if we denote our decoder as $\boldsymbol{\mu}^{t+1}_j=f_{\mathrm{dec}}(\bx^t_j)$ then we have:
\begin{align*}
\boldsymbol{\mu}^{2}_j&=f_{\mathrm{dec}}(\bx^1_j)& \\
\boldsymbol{\mu}^{t+1}_j&=f_{\mathrm{dec}}(\boldsymbol{\mu}^t_j)&  \quad t=2,\ldots,M \\
\boldsymbol{\mu}^{M+2}_j&=f_{\mathrm{dec}}(\bx^{M+1}_j)& \quad   \\
\boldsymbol{\mu}^{t+1}_j&=f_{\mathrm{dec}}(\boldsymbol{\mu}^t_j)&  \quad t=M+2,\ldots,2M \\
&\cdots &
\end{align*}
 We are backpropagating through this whole process, and since the errors accumulate for $M$ steps the degenerate decoder is now highly suboptimal.
\subsection{Recurrent decoder}
In many applications the Markovian assumption used in Sec.~\ref{sec:decoder} does not hold. To handle such applications we use a recurrent decoder that can model $p_\theta(\bx^{t+1}|\bx^t,...,\bx^1,\bz)$. Our recurrent decoder adds a GRU \cite{cho2014learning} unit to the GNN message passing operation. More formally:
\begin{align}
v{\rightarrow}e:\,\,\,\tilde{\bh}^{t}_{(i,j)} &= \sum_k z_{ij,k} \tilde{f}^k_{e}([\tilde{\bh}^t_i,\tilde{\bh}^t_j]) \label{eq:rec_dec_v2e}\\
e{\rightarrow}v:\,\,\,\mathrm{MSG}^{t}_j &= \textstyle \textstyle\sum_{i\neq j}\tilde{\bh}^{t}_{(i,j)}\label{eq:rec_dec_e2v}\\
\tilde{\bh}^{t+1}_j &= \mathrm{GRU}( [\mathrm{MSG}^{t}_j,\bx^{t}_j],\tilde{\bh}^{t}_j)\label{eq:rec_dec_??} \\
\boldsymbol{\mu}^{t+1}_j&=\bx^{t}_j+f_{\mathrm{out}}(\tilde{\bh}^{t+1}_j)\\
p(\bx^{t+1}|\bx^t,\bz) &= \mathcal{N}(\boldsymbol{\mu}^{t+1},\sigma^2 \mathbf{I})
\label{eq:rec_dec3}
\end{align}
The input to the message passing operation is the recurrent hidden state at the previous time step. $f_{\mathrm{out}}$ denotes an output transformation, modeled by a small MLP. For each node $v_j$ the input to the GRU update is the concatenation of the aggregated messages $\mathrm{MSG}^{t+1}_j$, the current input $\bx^{t+1}_j$, and the previous hidden state $\tilde{\bh}^{t}_j$.

If we wish to predict multiple time steps in the recurrent setting, the method suggested in Sec.~\ref{sec:degenerate} will be problematic. Feeding in the predicted (potentially incorrect) path and then periodically jumping back to the true path will generate artifacts in the learned trajectories. In order to avoid this issue we provide the correct input $\bx^t_j$ in the first $(T-M)$ steps, and only utilize our predicted mean $\boldsymbol{\mu}^t_j$ as input at the last $M$ time steps.
\subsection{Training}
Now that we have described all the elements, the training goes as follows: Given training example $\bx$ we first run the encoder and compute $q_\phi(\bz_{ij}|\bx)$, then we sample $\bz_{ij}$ from the concrete reparameterizable approximation of $q_\phi(\bz_{ij}|\bx)$. We then run the decoder to compute $\boldsymbol{\mu}^2,...,\boldsymbol{\mu}^{T}$. The ELBO objective, Eq.~\ref{eq:elbo}, has two terms: the reconstruction error $\mathbb{E}_{q_{\phi}(\bz|\bx)}[\log p_\theta(\bx|\bz)]$ and KL divergence $\mathrm{KL}[q_\phi(\bz|\bx)||p_\theta(\bz)]$. The reconstruction error is estimated by:
\begin{equation}
-\sum_j\sum_{t=2}^T\frac{||\bx^t_j-\boldsymbol{\mu}^t_j||^2}{2\sigma^2}+\mathrm{const}
\end{equation}
while the KL term for a uniform prior is just the sum of entropies (plus a constant):
\begin{equation}
\sum_{i\neq j}H(q_\phi(\bz_{ij}|\bx))+\mathrm{const}.
\end{equation}
As we use a reparameterizable approximation, we can compute gradients by backpropagation and optimize.
\section{Related Work}
Several recent works have studied the problem of learning the dynamics of a physical system from simulated trajectories \cite{BattagliaPLRK16,guttenberg2016permutation,chang2017compositional} and from generated video data \cite{WattersZWBPT17,steenkiste2018relational} with a graph neural network. Unlike our work they either assume a known graph structure or infer interactions implicitly.

Recent related works on graph-based methods for human motion prediction include \cite{AlahiGRRLS16} where the graph is not learned but is based on proximity and \cite{LeY0L17}  tries to cluster agents into roles.

A number of recent works \cite{monti2017geometric,duan2017one,hoshen2017vain,velickovic2018graph,garcia2018few,steenkiste2018relational} parameterize messages in GNNs with a soft attention mechanism \cite{luong2015effective,bahdanau2014neural}. This equips these models with the ability to focus on specific interactions with neighbors when aggregating messages. Our work is different from this line of research, as we explicitly perform inference over the latent graph structure. This allows for the incorporation of prior beliefs (such as sparsity) and for an interpretable discrete structure with multiple relation types.

\begin{figure}[tp]
\centering
  \includegraphics[width=\linewidth,trim={0 0 0 0.5cm},clip]{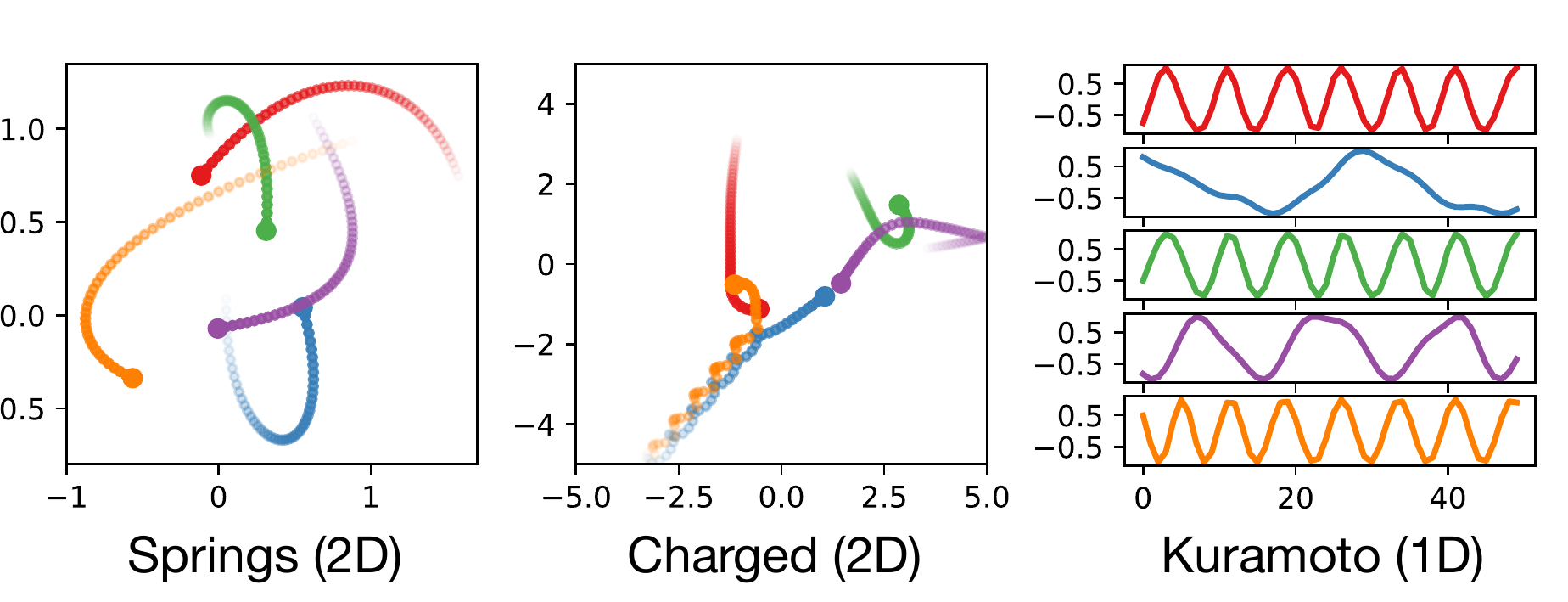}
  \vspace{-1em}
  \caption{Examples of trajectories used in our experiments from simulations of particles connected by springs (left), charged particles (middle), and phase-coupled oscillators (right).}
  \label{fig:simulators}
\end{figure}

The problem of inferring interactions or latent graph structure has been investigated in other settings in different fields. For example, in causal reasoning Granger causality \cite{Granger69} infers causal relations. Another example from computational neuroscience is   \cite{LindermanAP16,LindermanA14} where they infer interactions between neural spike trains.

\section{Experiments}
Our encoder implementation uses fully-connected networks (MLPs) or 1D CNNs with attentive pooling as our message passing function. For our decoder we used fully-connected networks or alternatively a recurrent decoder.  Optimization was performed using the Adam algorithm \cite{KingmaB14}. We provide full implementation details in the supplementary material. Our implementation uses PyTorch \cite{paszke2017automatic} and is available online\footnote{\url{https://github.com/ethanfetaya/nri}}.

\begin{figure*}[t!]
\centering
  \includegraphics[width=\textwidth]{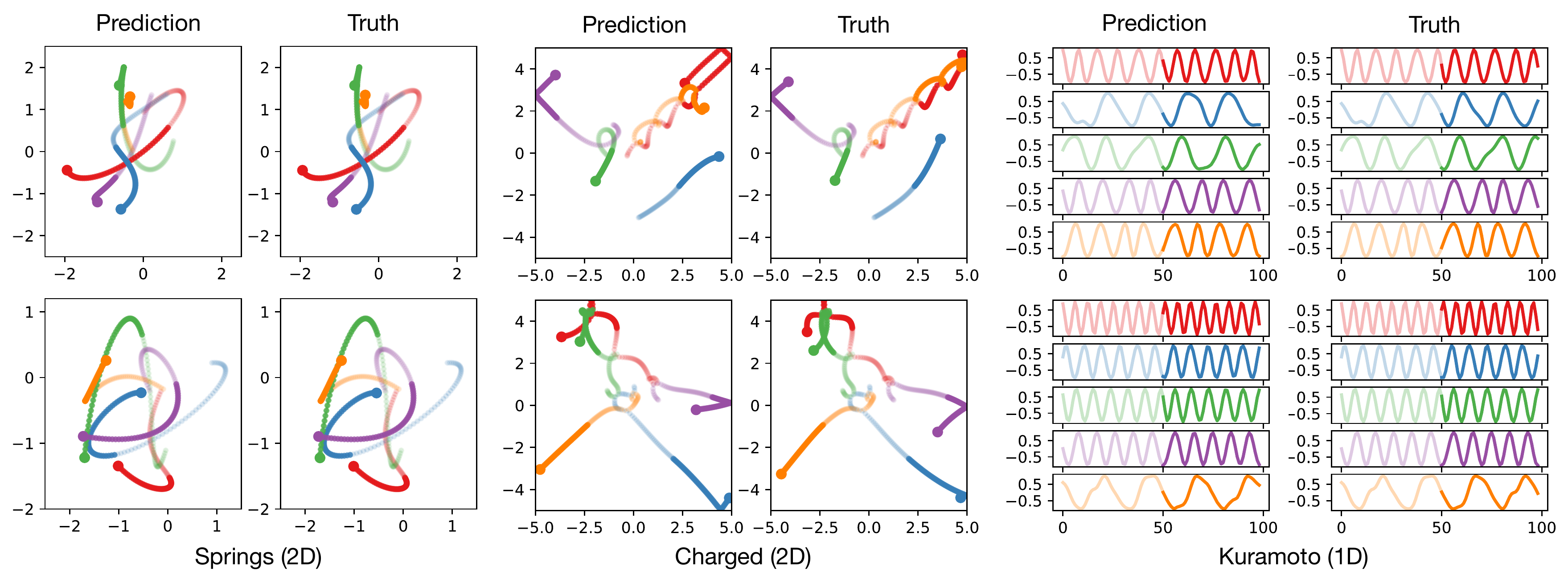}
  \caption{Trajectory predictions from a trained NRI model (unsupervised). Semi-transparent paths denote the first 49 time steps of ground-truth input to the model, from which the interaction graph is estimated. Solid paths denote self-conditioned model predictions.}
\label{fig:predictions}
\end{figure*}

\subsection{Physics simulations}\label{sec:phys_exp}
 We experimented with three simulated systems: particles connected by springs, charged particles and phase-coupled oscillators (Kuramoto model) \cite{Kuramoto75}. These settings allow us to attempt to learn the dynamics and interactions when the interactions are known. These systems, controlled by simple rules, can exhibit complex dynamics. For the springs and Kuramoto experiments the objects do or do not interact with equal probability. For the charged particles experiment they attract or repel with equal probability.  Example trajectories can be seen in Fig.~\ref{fig:simulators}. We generate 50k training examples, and 10k validation and test examples for all tasks. Further details on the data generation and implementation are in the supplementary material.

We note that the simulations are differentiable and so we can use it as a ground-truth decoder to train the encoder. The charged particles simulation, however, suffers from instability which led to some performance issues when calculating gradients; see supplementary material for further details. We used an external code base \cite{KuramotoImplementation} for stable integration of the Kuramoto ODE and therefore do not have access to gradient information in this particular simulation.

\paragraph{Results}

\begin{table}[t]
\centering
\vspace{-0.7em}
\caption{Accuracy (in $\%$) of unsupervised interaction recovery.}\label{tab:edge_acc} \vspace{0.8em}
\begin{tabular}{l c c c}
\toprule
\textbf{Model} & \textbf{Springs} & \textbf{Charged} & \textbf{Kuramoto} \\\midrule\\[-1.1em]
\multicolumn{4}{c}{5 objects}\\[0.1em]
Corr. (path)  & $52.4${\tiny$\pm$0.0} & $55.8${\tiny$\pm$0.0} & $62.8${\tiny$\pm$0.0}\\
Corr. (LSTM) & $52.7${\tiny$\pm$0.9} & $54.2${\tiny$\pm$2.0} & $54.4${\tiny$\pm$0.5}\\
NRI (sim.) & $\mathbf{99.8}${\tiny$\pm$0.0} & $59.6${\tiny$\pm$0.8} & --\\
NRI (learned) & $\mathbf{99.9}${\tiny$\pm$0.0} & $\mathbf{82.1}${\tiny$\pm$0.6} & $\mathbf{96.0}${\tiny$\pm$0.1}\\
\midrule\midrule
Supervised & $99.9${\tiny$\pm$0.0} & $95.0${\tiny$\pm$0.3} &  $99.7${\tiny$\pm$0.0}\\
\midrule\\[-1.1em]
\multicolumn{4}{c}{10 objects}\\[0.1em]
Corr. (path)  & $50.4${\tiny$\pm$0.0} & $51.4${\tiny$\pm$0.0} & $59.3${\tiny$\pm$0.0}\\
Corr. (LSTM) & $54.9${\tiny$\pm$1.0} & $52.7${\tiny$\pm$0.2} & $56.2${\tiny$\pm$0.7}\\
NRI (sim.) & $\mathbf{98.2}${\tiny$\pm$0.0} & $53.7${\tiny$\pm$0.8} & --\\
NRI (learned)  & $\mathbf{98.4}${\tiny$\pm$0.0} & $\mathbf{70.8}${\tiny$\pm$0.4}  & $\mathbf{75.7}${\tiny$\pm$0.3}\\
\midrule\midrule
Supervised & $98.8${\tiny$\pm$0.0} & $94.6${\tiny$\pm$0.2} & $97.1${\tiny$\pm$0.1}\\
\bottomrule
\end{tabular}
\vspace{-0.1em}
\end{table}

\begin{table*}[t!]
\centering
\vspace{-0.5em}
\caption{Mean squared error (MSE) in predicting future states for simulations with 5 interacting objects.}
\vspace{0.8em}
\begin{tabular}{l|ccc|ccc|ccc}
\toprule
& \multicolumn{3}{c}{\textbf{Springs}} & \multicolumn{3}{c}{\textbf{Charged}} & \multicolumn{3}{c}{\textbf{Kuramoto}} \\ \midrule
Prediction steps & 1 & 10 & 20 & 1 & 10 & 20 & 1 & 10 & 20  \\ \midrule
Static   & 7.93e-5 & 7.59e-3 & 2.82e-2 & 5.09e-3 & 2.26e-2 & 5.42e-2 & 5.75e-2 & 3.79e-1 & 3.39e-1\\
LSTM (single) & 2.27e-6 & 4.69e-4 & 4.90e-3 & 2.71e-3 & 7.05e-3 & 1.65e-2 & 7.81e-4 & 3.80e-2 & 8.08e-2\\
LSTM (joint)      & 4.13e-8 & 2.19e-5 & 7.02e-4 & 1.68e-3 & 6.45e-3 & 1.49e-2 & \textbf{3.44e-4} & \textbf{1.29e-2} & 4.74e-2\\
NRI (full graph)     & 1.66e-5 & 1.64e-3 & 6.31e-3 & \textbf{1.09e-3} & 3.78e-3 & 9.24e-3 & 2.15e-2 & 5.19e-2 & 8.96e-2 \\
NRI (learned)     & \textbf{3.12e-8} & \textbf{3.29e-6} & \textbf{2.13e-5} & \textbf{1.05e-3} & \textbf{3.21e-3} & \textbf{7.06e-3} & 1.40e-2 & 2.01e-2 & \textbf{3.26e-2}\\
\midrule
NRI (true graph)  & 1.69e-11 & 1.32e-9 & 7.06e-6 & 1.04e-3 & 3.03e-3 & 5.71e-3 & 1.35e-2 & 1.54e-2 & 2.19e-2\\
\bottomrule
\end{tabular}
\vspace{-0.5em}
\label{tab:pred_mse}
\end{table*}

We ran our NRI model on all three simulated physical  systems and compared our performance, both in future state prediction and in accuracy of estimating the edge type in an unsupervised manner.

For edge prediction, we compare to the ``gold standard" i.e.~training our encoder in a supervised way given the ground-truth labels. We also compare to the following baselines: Our NRI model with the ground-truth simulation decoder, NRI (sim.), and two correlation based baselines, Corr. (path) and Corr.~(LSTM). Corr.~(path) estimates the interaction graph by thresholding the matrix of correlations between trajectory feature vectors. Corr.~(LSTM) trains an LSTM \cite{hochreiter1997long} with shared parameters to model each trajectory individually and calculates correlations between the final hidden states to arrive at an interaction matrix after thresholding. We provide further details on these baselines in the supplementary material.

Results for the unsupervised interaction recovery task are summarized in Table \ref{tab:edge_acc} (average over 5 runs and standard error). As can be seen, the unsupervised NRI model, NRI (learned), greatly surpasses the baselines and recovers the ground-truth interaction graph with high accuracy on most tasks. For the springs model our unsupervised method is comparable to the supervised ``gold standard" benchmark. We note that our supervised baseline is similar to the work by \cite{SantoroRBMPBL17}, with the difference that we perform multiple rounds of message passing in the graph. Additional results on experiments with more than two edge types and non-interacting particles are described in the supplementary material.

For future state prediction we compare to the static baseline, i.e.~$\bx^{t+1}=\bx^t$, two LSTM baselines, and a full graph baseline. One LSTM baseline, marked as ``single", runs a separate LSTM (with shared weights) for each object. The second, marked as ``joint" concatenates all state vectors and feeds it into one LSTM that is trained to predict all future states simultaneously. Note that the latter will only be able to operate on a fixed number of objects (in contrast to the other models).

In the full graph baseline, we use our  message passing decoder on the fully-connected graph without edge types, i.e.~without inferring edges. This is similar to the model used in \cite{WattersZWBPT17}. We also compare to the ``gold standard" model, denoted as NRI (true graph), which is training only a decoder using the ground-truth graph as input. The latter baseline is comparable to previous works such as interaction networks \cite{BattagliaPLRK16}.

In order to have a fair comparison, we generate longer test trajectories and only evaluate on the last part unseen by the encoder. Specifically, we run the encoder on the first 49 time steps (same as in training and validation), then predict with our decoder the following 20 unseen time steps. For the LSTM baselines, we first have a ``burn-in" phase where we feed the LSTM the first 49 time steps, and then predict the next 20 time steps. This way both algorithms have access to the first 49 steps when predicting the next 20 steps. We show mean squared error (MSE) results in Table \ref{tab:pred_mse}, and note that our results are better than using LSTM for long term prediction. Example trajectories predicted by our NRI (learned) model for up to 50 time steps are shown in Fig.~\ref{fig:predictions}.

\begin{figure}[tp]
  \centering
      \begin{subfigure}[b]{0.495\linewidth}
  \includegraphics[width=\linewidth,trim={0.25cm 0 0.35cm 0},clip]{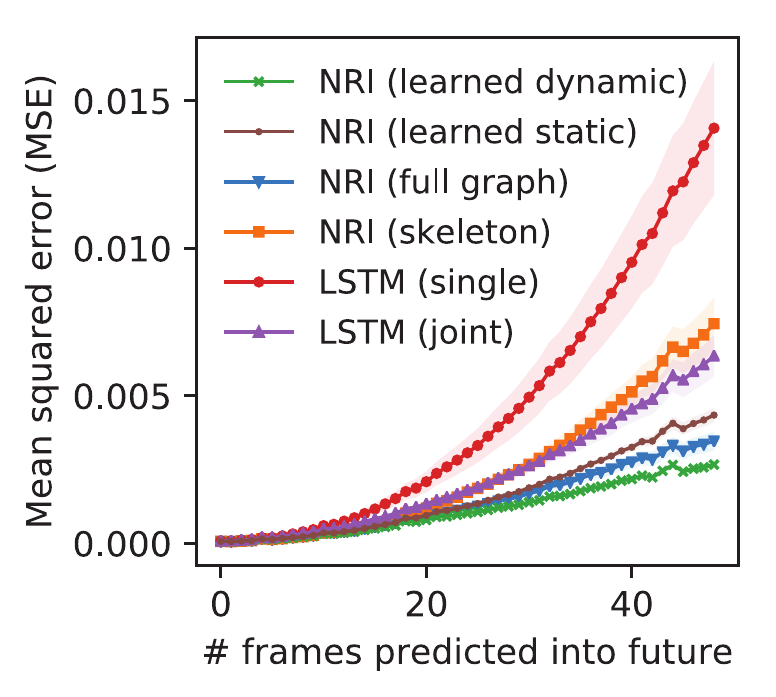}
  \end{subfigure}
  ~
  \begin{subfigure}[b]{0.465\linewidth}
  \includegraphics[width=\linewidth,trim={0.25cm 0 0.35cm 0},clip]{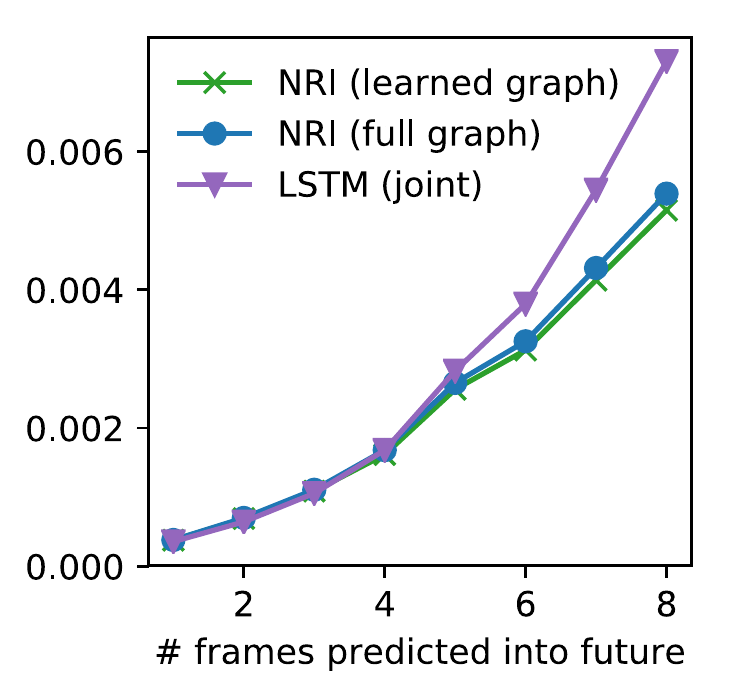}
  \end{subfigure}
  \vspace{-1em}
  \caption{Test MSE comparison for motion capture (walking) data (\textit{left}) and sports tracking (SportVU) data (\textit{right}).}
  \label{fig:mocap}
\end{figure}

For the Kuramoto model, we observe that the LSTM baselines excel at smoothly continuing the shape of the waveform for short time frames, but fail to model the long-term dynamics of the interacting system. We provide further qualitative analysis for these results in the supplementary material.

It is interesting to note that the charged particles experiment achieves an MSE score which is on par with the NRI model given the true graph, while only predicting 82.6\% of the edges accurately. This is explained by the fact that far away  particles have weak interactions, which have only small effects on future prediction. An example can be seen in Fig. \ref{fig:predictions} in the top row where the blue particle is repelled instead of being attracted.

\subsection{Motion capture data}

The CMU Motion Capture Database \cite{cmu2003mocap} is a large collection of motion capture recordings for various tasks (such as walking, running, and  dancing) performed by human subjects. We here focus on recorded walking motion data of a single subject (subject \#35). The data is in the form of 31 3D trajectories, each tracking a single joint. We split the different walking trials into non-overlapping training (11 trials), validation (4 trials) and test sets (7 trials). We provide both position and velocity data. See supplementary material for further details. We train our NRI model with an MLP encoder and RNN decoder on this data using 2 or 4 edge types where one edge type is ``hard-coded" as non-edge, i.e.~messages are only passed on the other edge types. We found that experiments with 2 and 4 edge types give almost identical results, with two edge types being comparable in capacity to the fully connected graph baseline while four edge types (with sparsity prior) are more interpretable and allow for easier visualization.

\begin{figure}[tp]
  \centering
    \begin{subfigure}[b]{0.43\linewidth}
        \includegraphics[width=\textwidth]{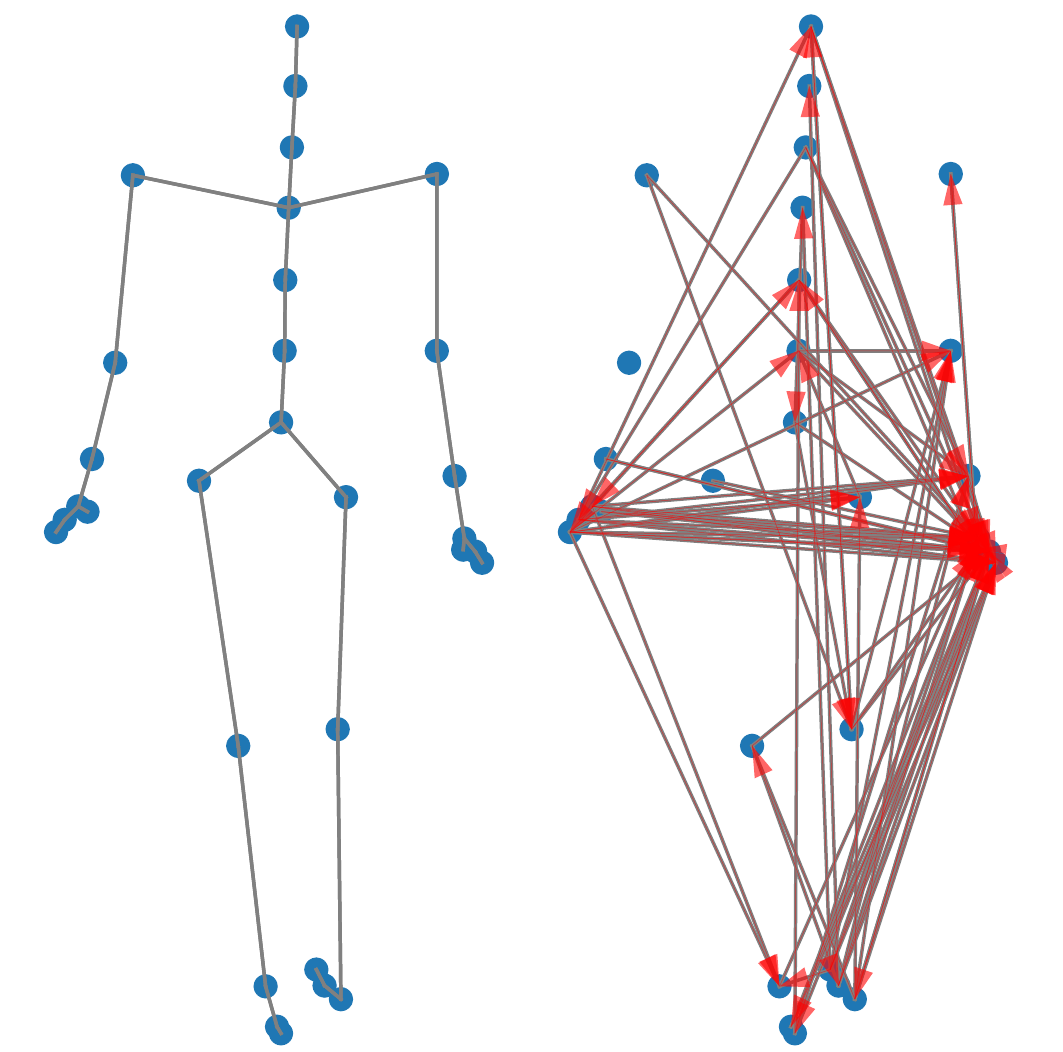}
        \caption{Right hand focus}
        \label{fig:mocap_graph1}
    \end{subfigure}
    ~
    \begin{subfigure}[b]{0.43\linewidth}
        \includegraphics[width=\textwidth]{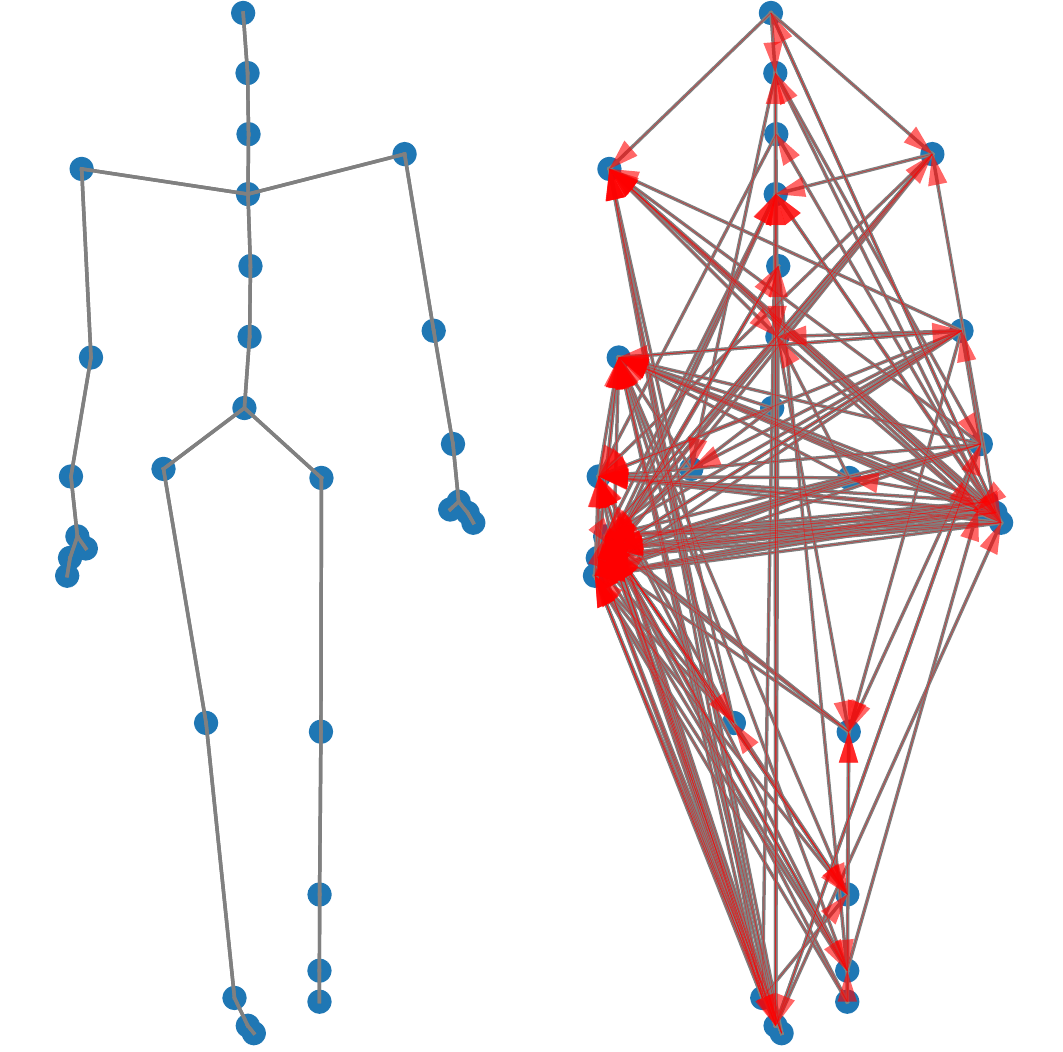}
        \caption{Left hand focus}
        \label{fig:mocap_graph2}
    \end{subfigure}
  \vspace{-1em}
  \caption{Learned latent graphs on motion capture data (4 edge types)\footnotemark. Skeleton shown for reference. Red arrowheads denote directionality of a learned edge. The edge type shown favors a specific hand depending on the state of the movement and gathers information mostly from other extremities.}
  \label{fig:mocap_graph}
\end{figure}

\paragraph{Dynamic graph re-evaluation}
We find that the learned graph depends on the particular phase of the motion (Fig.~\ref{fig:mocap_graph}), which indicates that the ideal underlying graph is dynamic. To account for this, we dynamically re-evaluate the NRI encoder for every time step during testing, effectively resulting in a dynamically changing latent graph that the decoder can utilize for more accurate predictions.

\paragraph{Results}
The qualitative results for our method and the same baselines used in Sec. \ref{sec:phys_exp} can be seen in Fig.~\ref{fig:mocap}. As one can see, we outperform the fully-connected graph setting in long-term predictions, and both models outperform the LSTM baselines. Dynamic graph re-evaluation significantly improves predictive performance for this dataset compared to a static baseline. One interesting observation is that the skeleton graph is quite suboptimal, which is surprising as the skeleton is the ``natural'' graph. When examining the edges found by our model (trained with 4 edge types and a sparsity prior) we see an edge type that mostly connects a hand to other extremities, especially the opposite hand, as seen in Fig.~\ref{fig:mocap_graph}. This can seem counter-intuitive as one might assume that the important connections are local, however we note that some leading approaches for modeling motion capture data \cite{jain2016structural} do indeed include hand to hand interactions.

\footnotetext{The first edge type is ``hard-coded" as non-edge and was trained with a prior probability of 0.91. All other edge types received a prior of 0.03 to favor sparse graphs that are easier to visualize. We visualize test data not seen during training.}

\subsection{Pick and Roll NBA data}
The National Basketball Association (NBA) uses the SportVU
tracking system to collect player tracking data, where each frame contains the location of all ten players and the ball. Similar to our previous experiments, we test our model on the task of future trajectory prediction. Since the interactions between players are dynamic, and our current formulation assumes fixed interactions during training, we focus on the short Pick and Roll (PnR) instances of the games.  PnR is one of the most common offensive tactics in the NBA where an offensive player sets a screen for the ball handler, attempting to create separation between the ball handler and his matchup.

We extracted 12k segments from the 2016 season and used 10k, 1k, 1k for training, validation, and testing respectively.  The segments are 25 frames long (i.e.~4 seconds) and consist of only 5 nodes: the ball, ball hander, screener, and defensive matchup for each of the players.
\vspace{1em}

We trained a CNN encoder and a RNN decoder with 2 edge types. For fair comparison, and because the trajectory continuation is not PnR anymore, the encoder is trained on only the first 17 time steps (as deployed in testing). Further details are in the supplementary material. Results for test MSE are shown in Figure \ref{fig:mocap}. Our model outperforms a baseline LSTM model, and is on par with the full graph.

\begin{figure}[tp]
  \centering
  \includegraphics[width=0.98\linewidth]{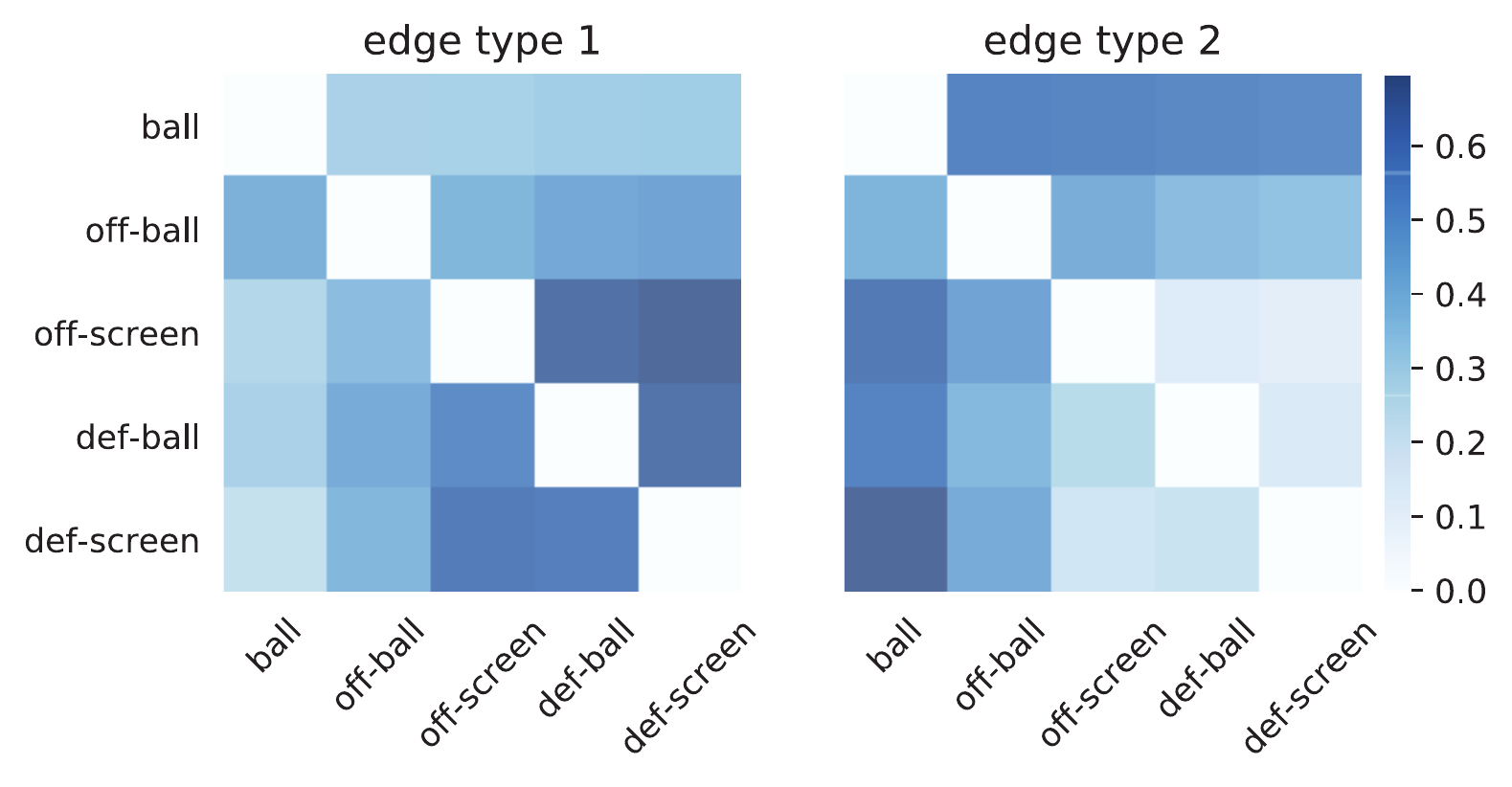}
  \vspace{-1em}
  \caption{Distribution of learned edges between players (and the ball) in the basketball sports tracking (SportVU) data.}
  \label{fig:sportvu}
\end{figure}

To understand the latent edge types we show in Fig. \ref{fig:sportvu} how they are distributed between the players and the ball. As we can see, one edge type mostly connects ball and ball handler (off-ball) to all other players, while the other is mostly inner connections between the other three players. As the ball and ball handler are the key elements in the PnR play, we see that our model does learn an important semantic structure by separating them from the rest.

\section{Conclusion}
In this work we introduced NRI, a method to simultaneously infer relational structure while learning the dynamical model of an interacting system. In a range of experiments with physical simulations we demonstrate that our NRI model is highly effective at unsupervised recovery of ground-truth interaction graphs. We further found that it can model the dynamics of interacting physical systems, of real motion tracking and of sports analytics data at a high precision, while learning reasonably interpretable edge types.

Many real-world examples, in particular multi-agent systems such as traffic, can be understood as an interacting system where the interactions are dynamic.  While our model is trained to discover static interaction graphs, we demonstrate that it is possible to apply a trained NRI model to this evolving case by dynamically re-estimating the latent graph. Nonetheless, our solution is limited to static graphs during training and future work will investigate an extension of the NRI model that can explicitly account for dynamic latent interactions even at training time.

\section*{Acknowledgements}
The authors would like to thank the Toronto Raptors and the NBA for the use of the SportVU data. We would further like to thank Christos Louizos and Elise van der Pol for helpful discussions. This project is supported by the SAP Innovation Center Network.

\bibliography{references}
\bibliographystyle{icml2018}

\newpage

\appendix
\section{Further experimental analysis}

\subsection{Kuramoto LSTM vs. NRI comparison}
\begin{figure*}[htp!]
  \centering
  \includegraphics[width=0.7\textwidth]{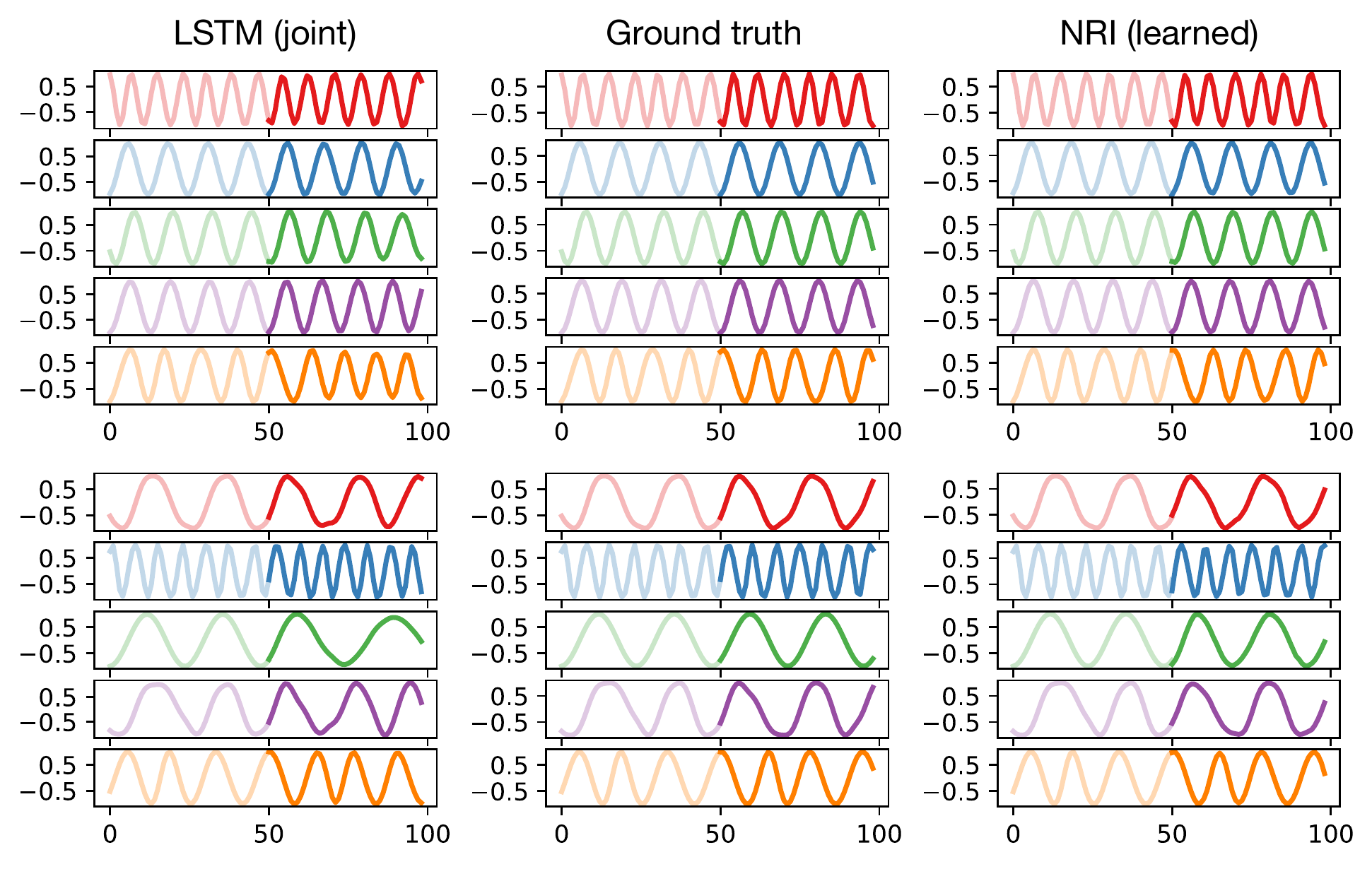}
  \caption{Qualitative comparison of model predictions for the LSTM (joint) model (\textit{left}) and the NRI (learned) model (\textit{right}). The ground truth trajectories (\textit{middle}) are shown for reference.}
  \label{fig:lstm_nri_kurmaoto}
\end{figure*}

From the results in our main paper it became evident that a simple LSTM model excels at predicting the dynamics of a network of phase-coupled oscillators (Kuramoto model) for short periods of time, while predictive performance deteriorates for longer sequences. It is interesting to compare the qualitative predictive behavior of this fully recurrent model with our NRI (learned) model that models the state $\mathbf{x}^{t+1}$ at time $t+1$ solely based on the state $\mathbf{x}^{t}$ at time t and the learned latent interaction graph. In Fig.~\ref{fig:lstm_nri_kurmaoto} we provide visualizations of model predictions for the LSTM (joint) and the NRI (learned) model, compared to the ground truth continuation of the simulation.

It can be seen that the LSTM model correctly captures the shape of the sinusoidal waveform but fails to model the phase dynamics that arise due to the interactions between the oscillators. Our NRI (learned) model captures the qualitative behavior of the original coupled model at a very high precision and only in some cases slightly misses the phase dynamics (e.g.~in the purple and green curve in the lower right plot). The LSTM model rarely matches the phase of the ground truth trajectory in the last few time steps and often completely goes ``out of sync'' by up to half a wavelength.

\subsection{Spring simulation variants}
In addition to the experiments presented in the main paper, we analyze the following two variants of the spring simulation experimental setting: i) we test a trained model on completely non-interacting (free-floating) particles, and ii) we add a third edge type with a lower coupling constant.

To test whether our model can infer an empty graph, we create a test set of 1000 simulations with 5 non-interacting particles and test an unsupervised NRI model which was trained on the spring simulation dataset with 5 particles as before. We find that it achieves an accuracy of $98.4\%$ in identifying "no interaction" edges (i.e. the empty graph).

The last variant explores a simulation with more than two known edge types. We follow the same procedure for the spring simulation with 5 particles as before with the exception of adding an additional edge type with coupling constant $k_{ij}=0.5$ (all three edge types are sampled with equal probability). We fit an unsupervised NRI model to this data ($K=3$ in this case, other settings as before) and find that it achieves an accuracy of $99.2\%$ in discovering the correct edge types.

\subsection{Motion capture visualizations}
\begin{figure*}[htp]
  \centering
  \begin{subfigure}[b]{0.75\linewidth}
  \includegraphics[width=\textwidth]{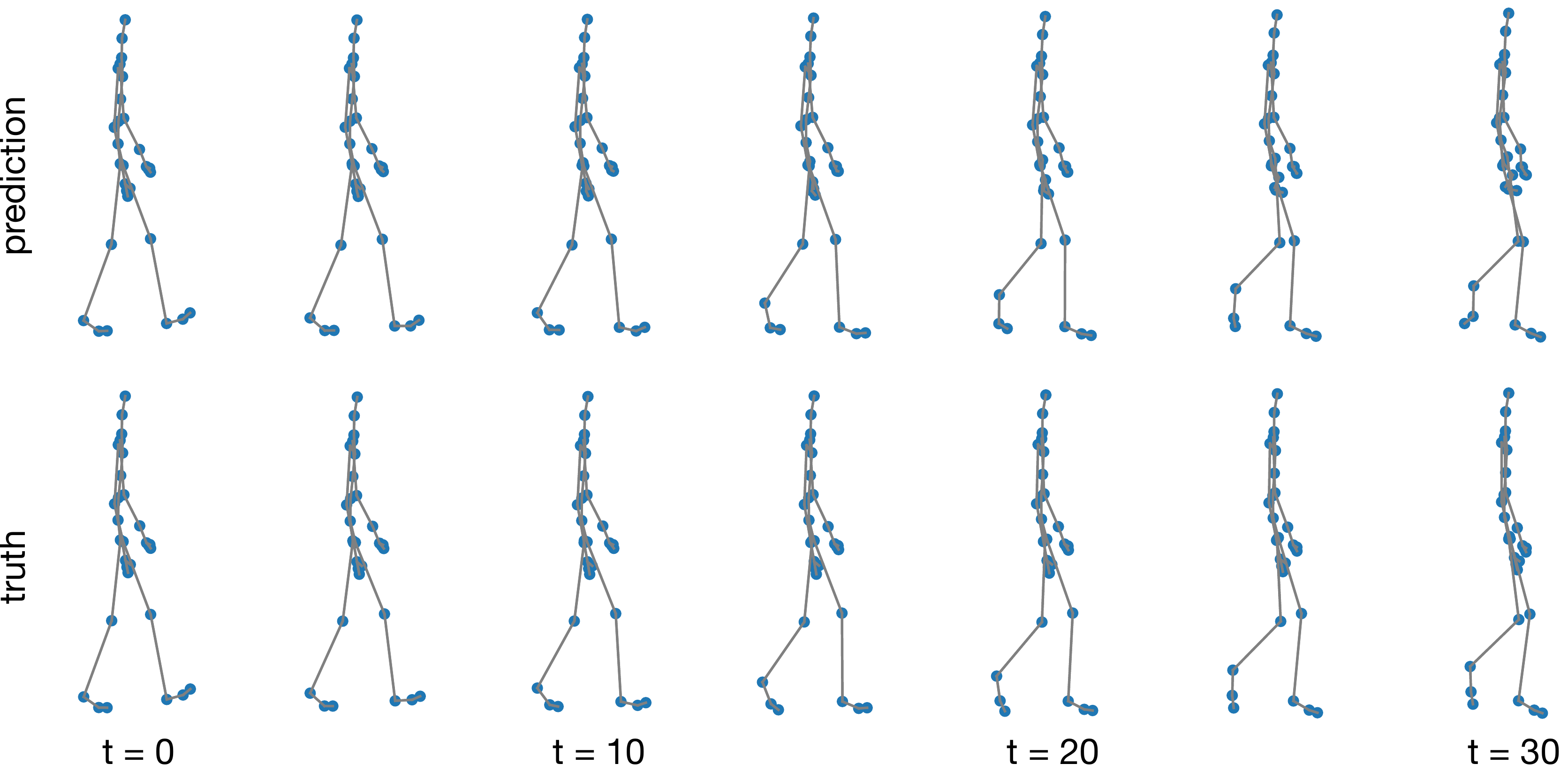}
    \caption{Test trial 1.}
     ~\vspace{1em}
  \end{subfigure}
  \begin{subfigure}[b]{0.75\linewidth}
  \includegraphics[width=\textwidth]{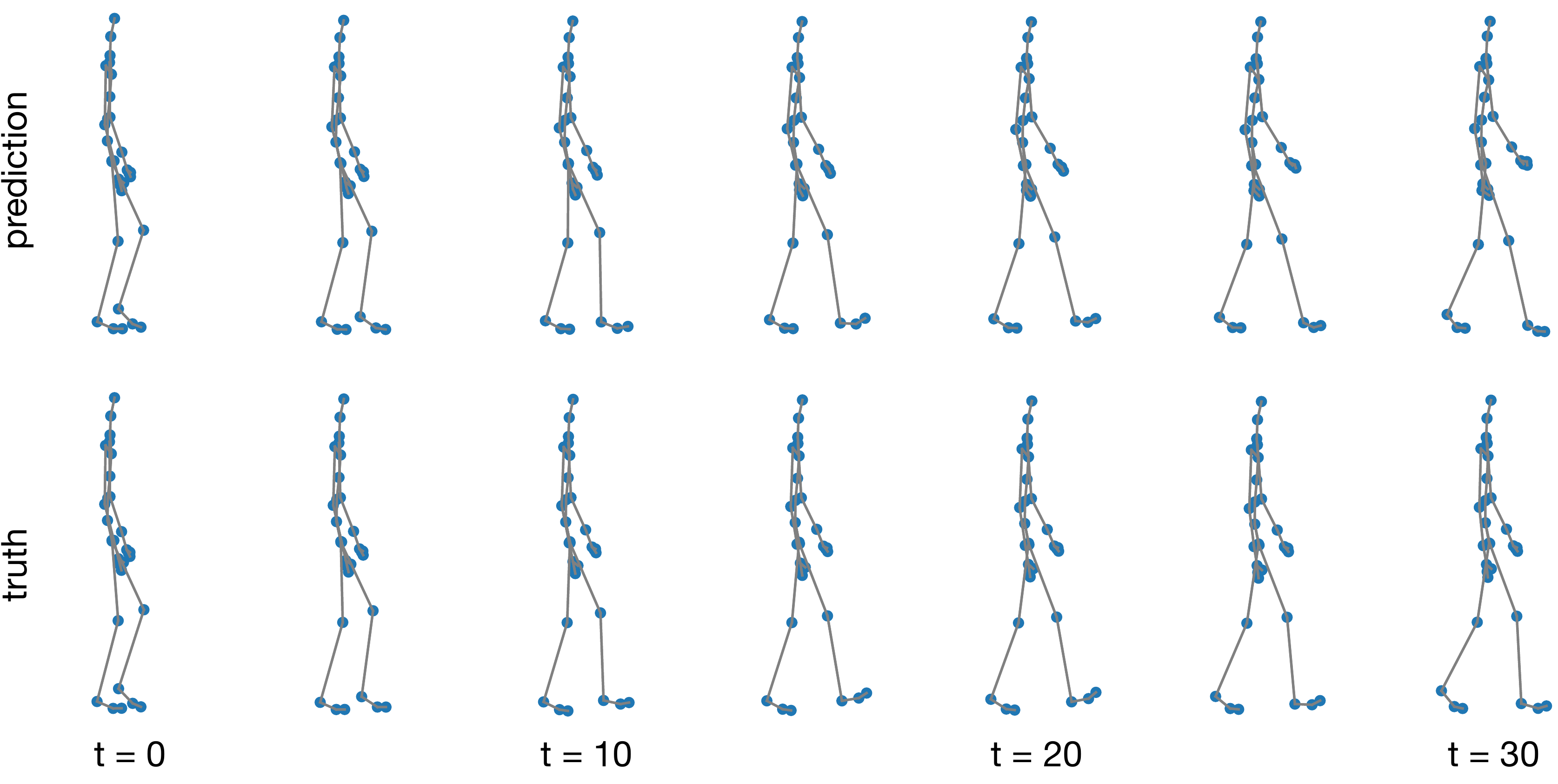}
      \caption{Test trial 2.}
  \end{subfigure}
  \caption{Examples of predicted walking motion of an NRI model with learned latent graph compared to ground truth sequences for two different test set trials.}
  \label{fig:mocap_walk}
\end{figure*}

In Fig.~\ref{fig:mocap_walk} we visualize predictions of a trained NRI model with learned latent graph for the motion capture dataset. We show 30 predicted time steps of future movement, conditioned on 49 time steps that are provided as ground truth to the model. It can be seen that the model can capture the overall form of the movement with high precision. Mistakes (e.g.~the misplaced toe node in frame 30) are possible due to the accumulation of small errors when predicting over long sequences with little chance of recovery. Curriculum learning schemes where noise is gradually added to training sequences can potentially alleviate this issue.

\subsection{NBA visualizations}
\begin{figure*}[htp]
  \includegraphics[width=\textwidth]{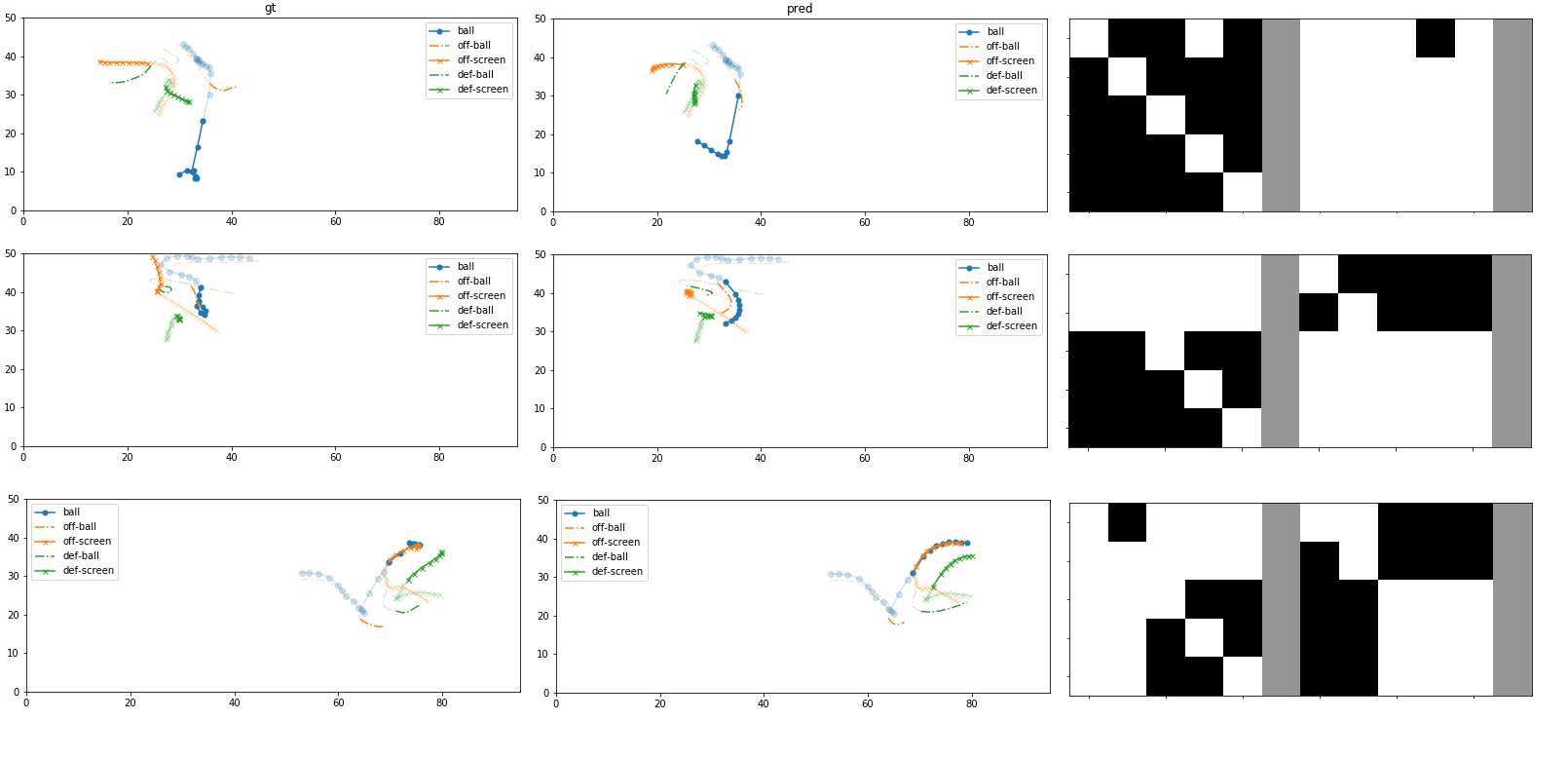}
    \caption{Visualization of NBA trajectories. \textit{Left}: ground truth; \textit{middle}: model prediction; \textit{right}: sampled edges.}
      \label{fig:NBA}

\end{figure*}

We show examples of three pick and roll trajectories in Fig. \ref{fig:NBA}. In the left column we show the ground truth, in the middle we show our prediction and in the right we show the edges that where sampled by our encoder. As we can see even when our model does not predict the true future path, which is extremely challenging for this data, it still makes semantically reasonable predictions.  For example in the middle row it predicts that the player defending the ball handler passes between him and the screener (going over the screen) which is a reasonable outcome even though in reality the defenders switched players.

\section{Simulation data}
\subsection{Springs model}
We simulate $N\in\{5,10\}$ particles (point masses) in a 2D box with no external forces (besides elastic collisions with the box). We randomly connect, with probability $0.5$, each pair of particles with a spring. The particles connected by springs interact via forces given by Hooke's law $F_{ij}=-k(r_i-r_j)$ where $F_{ij}$ is the force applied to particle $v_i$ by particle $v_j$, $k$ is the spring constant and $r_{i}$ is the 2D location vector of particle $v_i$. The initial location is sampled from a Gaussian $\mathcal{N}(0,0.5)$, and the initial velocity is a random vector of norm $0.5$. Given the initial locations and velocity we can simulate the trajectories by solving Newton's equations of motion PDE. We do this by leapfrog integration using a step size of 0.001 and then subsample each 100 steps to get our training and testing trajectories.

We note that since the leapfrog integration is differentiable, we are able to use it as a ground-truth decoder and back-propagate through it to train the encoder. We implemented the leapfrog integration in PyTorch, which allows us to compare model performance with a learned decoder versus the ground-truth simulation decoder.

\subsection{Charged particles model}
Similar to the springs model, we simulate $N\in\{5,10\}$  particles in a 2D box, but instead of springs now our particles carry positive or negative charges $q_i\in\{\pm q\}$, sampled with uniform probability, and interact via Coulomb forces: $F_{ij}=C\cdot \textrm{sign}({q_i\cdot q_j})\frac{r_i-r_j}{||r_i-r_j||^3}$ where $C$ is some constant. Unlike the springs simulations, here every two particles interact, although the interaction might be weak if they stay far apart, but they can either attract or repel each other.

Since the forces diverge when the distance between particles goes to zero, this can cause issues when integrating with a fixed step size. The problem might be solved by using a much smaller step size, but this would slow the generation considerably. To circumvent this problem, we clip the forces to some maximum absolute value. While not being exactly physically accurate, the trajectories are indistinguishable to a human observer and the generation process is now stable.

The force clipping does, however, create a problem for the simulation ground-truth decoder, as gradients become zero when the forces are clipped during the simulation. We attempted to fix this by using ``soft" clipping with a $\mathrm{softplus}(x)=\log(1+e^x)$ function in the differentiable simulation decoder, but this similarly resulted in vanishing gradients once the model gets stuck in an unfavorable regime with large forces.

\subsection{Phase-coupled oscillators}
The Kuramoto model is a nonlinear system of phase-coupled oscillators that can exhibit a range of complicated dynamics based on the distribution of the oscillators' internal frequencies and their coupling strengths. We use the common form for the Kuramoto model given by the following differential equation:
\begin{align}
\frac{d\phi_i}{dt} = \omega_i + \sum_{j\neq i}k_{ij}\sin(\phi_i - \phi_j)
\label{eq:kuramoto}
\end{align}
with phases $\phi_i$, coupling constants $k_{ij}$, and intrinsic frequencies $\omega_i$. We simulate 1D trajectories by solving Eq.~\eqref{eq:kuramoto} with a fourth-order Runge-Kutta integrator with step size $0.01$.

We simulate $N\in\{5,10\}$ phase-coupled oscillators in 1D with intrinsic frequencies $\omega_i$ and initial phases $\phi_i^{t=1}$ sampled uniformly from $[1, 10)$ and  $[0, 2\pi)$, respectively. We randomly, with probability of $0.5$, connect pairs of oscillators $v_i$ and $v_j$ (undirected) with a coupling constant $k_{ij}=1$. All other coupling constants are set to 0. We subsample the simulated $\phi_i$ by a factor of 10 and create trajectories $\bx_i$ by concatenating $\frac{d\phi_i}{dt}$, $\sin{\phi_i}$, and the intrinsic frequencies $\omega_i$ (copied for every time step as $\omega_i$ are static).

\section{Implementation details}
We will describe here the details of our encoder and decoder implementations.
\subsection{Vectorized implementation}\label{sec:vec}
The message passing operations $v{\rightarrow}e$ and $v{\rightarrow}e$ can be evaluated in parallel for all nodes (or edges) in the graph and allow for an efficient vectorized implementation. More specifically, the node-to-edge message passing function $f_{v{\rightarrow}e}$ can be vectorized as:
\begin{equation}
\mathbf{H}^1_e = f_e([\mathbf{M}^{\mathrm{in}}_{v\rightarrow e}\mathbf{H}^1_v, \mathbf{M}^{\mathrm{out}}_{v\rightarrow e}\mathbf{H}^1_v])
\end{equation}
with $\mathbf{H}_v=[\mathbf{h}^\top_1, \mathbf{h}^\top_2, \ldots, \mathbf{h}^\top_N]^\top\in\mathbb{R}^{N\times F}$ and $\mathbf{H}_e\in\mathbb{R}^{E\times F}$ defined analogously (layer index omitted), where $F$ and $E$ are the total number of features and edges, respectively. $(\cdot)^\top$ denotes transposition. Both message passing matrices $\mathbf{M}_{v\rightarrow e}\in\mathbb{R}^{E\times N}$ are dependent on the graph structure and can be computed in advance if the underlying graph is static. $\mathbf{M}^{\mathrm{in}}_{v\rightarrow e}$ is a sparse binary matrix with $\mathbf{M}^{\mathrm{in}}_{v\rightarrow e, ij}=1$ when the $j$-th node is connected to the $i$-th edge (arbitrary ordering) via an incoming link and $0$ otherwise. $\mathbf{M}^{\mathrm{out}}_{v\rightarrow e}$ is defined analogously for outgoing edges.

Similarly, we can vectorize the edge-to-node message passing function $f_{e{\rightarrow}v}$ as:
\begin{equation}
\mathbf{H}^2_v = f_v(\mathbf{M}^{\mathrm{in}}_{e\rightarrow v}\mathbf{H}^1_e)
\end{equation}
with $\mathbf{M}^{\mathrm{in}}_{e\rightarrow v} = (\mathbf{M}^{\mathrm{in}}_{v\rightarrow e})^\top$. For large sparse graphs (e.g.~by constraining interactions to nearest neighbors), it can be beneficial to make use of sparse-dense matrix multiplications, effectively allowing for an $O(E)$ algorithm.

\subsection{MLP Encoder}\label{sec:mlp_encoder}
The basic building block of our MLP encoder is a 2-layer MLP with hidden and output dimension of 256, with batch normalization, dropout, and ELU activations. Given this, the forward model for our encoder is given by the code snippet in Fig.~\ref{code:enc_mlp}. The node2edge module returns for each edge the concatenation of the receiver and sender features. The edge2node module accumulates all incoming edge features via a sum.
\begin{figure}[htp!]
\begin{lstlisting}[language=python]
x = self.mlp1(x) # 2-layer ELU net per node
x = self.node2edge(x)
x = self.mlp2(x)
x_skip = x

x = self.edge2node(x)
x = self.mlp3(x)
x = self.node2edge(x)
x = torch.cat((x, x_skip), dim=2)
x = self.mlp4(x)
return self.fully_connected_out(x)
\end{lstlisting}
\vspace{-0.8em}
\caption{PyTorch code snippet of the MLP encoder forward pass.}\label{code:enc_mlp}
\end{figure}
\subsection{CNN Encoder}\label{sec:cnn_encoder}
The CNN encoder uses another block which performs 1D convolutions with attention. This allows for encoding with changing trajectory size, and is also appropriate for tasks like the charged particle simulations when the interaction can be strong for a small fraction of time. The forward computation of this module is presented in Fig.~\ref{code:cnn_block} and the overall decoder in Fig.~\ref{code:cnn_encoder}.
\begin{figure}[htp!]
\begin{lstlisting}[language=python]
# CNN block
# inputs is of shape ExFxT, E: number of edges,
# 	T: sequence length, F: num. features
x = F.relu(self.conv1(inputs))
x = self.batch_norm1(x)
x = self.pool(x)
x = F.relu(self.conv2(x))
x = self.batch_norm2(x)
out = self.conv_out(x)
attention = softmax(self.conv_attn(x), axis=2)

out = (out * attention).mean(dim=2)
return out
\end{lstlisting}
\vspace{-0.8em}
\caption{PyTorch code snippet of the CNN block forward pass, used in the CNN encoder.}\label{code:cnn_block}
\end{figure}

\begin{figure}[htp!]
\begin{lstlisting}[language=python]
# CNN encoder
x = self.node2edge(x)
x = self.cnn(x)  # CNN block from above
x = self.mlp1(x)  # 2-layer ELU net per node
x_skip = x

x = self.edge2node(x)
x = self.mlp2(x)
x = self.node2edge(x)
x = torch.cat((x, x_skip), dim=2)
x = self.mlp3(x)
return self.fully_connected_out(x)
\end{lstlisting}
\vspace{-0.8em}
\caption{PyTorch code snippet of the  CNN encoder model forward pass.}\label{code:cnn_encoder}
\end{figure}

\subsection{MLP Decoder}
In Fig. \ref{code:dec_mlp_step} we present the code for a single time-step prediction using our MLP decoder for Markovian data.
\begin{figure}[htp!]
\begin{lstlisting}[language=python]
# Single prediction step
pre_msg = self.node2edge(inputs)

# Run separate MLP for every edge type
# For non-edge: start_idx=1, otherwise 0
for i in range(start_idx, num_edges):
  msg = F.relu(self.msg_fc1[i](pre_msg))
  msg = F.relu(self.msg_fc2[i](msg))
  msg = msg * edge_type[:, :, :, i:i + 1]
  all_msgs += msg

# Aggregate all msgs to receiver
agg_msgs = self.edge2node(all_msgs)
hidden = torch.cat([inputs, agg_msgs], dim=-1)

# Output MLP
pred = F.relu(self.out_fc1(hidden)
pred = F.relu(self.out_fc2(pred)
pred = self.out_fc3(pred)

return inputs + pred
\end{lstlisting}
\vspace{-0.8em}
\caption{PyTorch code snippet of a single prediction step in the MLP decoder.}\label{code:dec_mlp_step}
\end{figure}

\subsection{RNN Decoder}
The RNN decoder adds a GRU style update to the single step prediction, the code snippet for the GRU module is presented in Fig.~\ref{code:dec_rnn_gru}  and the overall RNN decoder in Fig.~\ref{code:dec_rnn_step}.
\begin{figure}[htp!]
\begin{lstlisting}[language=python]
# GRU block
# Takes arguments: inputs, agg_msgs, hidden
r = F.sigmoid(self.input_r(inputs) +
	self.hidden_r(agg_msgs))
i = F.sigmoid(self.input_i(inputs) +
	self.hidden_i(agg_msgs))
n = F.tanh(self.input_n(inputs) +
	r * self.hidden_h(agg_msgs))
hidden = (1 - i) * n + i * hidden
return hidden
\end{lstlisting}
\vspace{-0.8em}
\caption{PyTorch code snippet of a GRU block, used in the RNN decoder.}\label{code:dec_rnn_gru}
\end{figure}

\begin{figure}[htp!]
\begin{lstlisting}[language=python]
# Single prediction step
pre_msg = self.node2edge(inputs)

# Run separate MLP for every edge type
# For non-edge: start_idx=1, otherwise 0
for i in range(start_idx, num_edges):
  msg = F.relu(self.msg_fc1[i](pre_msg))
  msg = F.relu(self.msg_fc2[i](msg))
  msg = msg * edge_type[:, :, :, i:i + 1]
  # Average over types for stability
  all_msgs += msg/(num_edges-start_idx)

# Aggregate all msgs to receiver
agg_msgs = self.edge2node(all_msgs)

# GRU-style gated aggregation (see GRU block)
hidden = self.gru(inputs, agg_msgs, hidden)

# Output MLP
pred = F.relu(self.out_fc1(hidden))
pred = F.relu(self.out_fc2(pred))
pred = self.out_fc3(pred)

# Predict position/velocity difference
pred = inputs + pred

return pred, hidden
\end{lstlisting}
\vspace{-0.8em}
\caption{PyTorch code snippet of a single prediction step in the RNN decoder.}\label{code:dec_rnn_step}
\end{figure}

\section{Experiment details}
All experiments were run using the Adam optimizer (Kingma \& Ba, 2015) with a learning rate of 0.0005, decayed by a factor of 0.5 every 200 epochs. Unless otherwise noted, we train with a batch size of 128. The concrete distribution is used with $\tau=0.5$. During testing, we replace the concrete distribution with a categorical distribution to obtain discrete latent edge types. Physical simulation and sports tracking experiments were run for 500 training epochs. For motion capture data we used 200 training epochs, as models tended to converge earlier. We saved model checkpoints after every epoch whenever the validation set performance (measured by path prediction MSE) improved and loaded the best performing model for test set evaluation. We observed that using significantly higher learning rates than 0.0005 often produced suboptimal decoders that ignored the latent graph structure.
\subsection{Physics simulations experiments}
The springs, charged particles and Kuramoto datasets each contain 50k training instances and 10k validation and test instances. Training and validation trajectories where of length 49 while test trajectories continue for another 20 time steps (50 for visualization). We train an MLP encoder for the springs experiment, and CNN encoder for the charged particles and Kuramoto experiments. All experiments used MLP decoders and two edge types. For the Kuramoto model experiments, we explicitly hard-coded the first edge type as a ``non-edge", i.e.~no messages are passed along edges of this type.

As noted previously, all of our MLPs have hidden and output dimension of 256. The overall input/output dimension of our model is 4 for the springs and charged particles experiments (2D position and velocity) and 3 for the Kuramoto model experiments (phase-difference, amplitude and intrinsic frequency). During training, we use teacher forcing in every 10-th time step (i.e. every 10th time step, the model receives a ground truth input, otherwise it receives its previous prediction as input). As we always have two edge types in these experiments and their ordering is arbitrary (apart from the Kuramoto model where we assign a special role to edge type 1), we choose the ordering for which the accuracy is highest.

\subsubsection{Baselines}
\paragraph{Edge recovery experiments}
In edge recovery experiments, we report the following baselines along with the performance of our NRI (learned) model:
\begin{itemize}
\item \textbf{Corr.~(path)}: We calculate a correlation matrix $R$, where $R_{ij}=\frac{C_{ij}}{\sqrt{C_{ii}C_{jj}}}$ with $C_{ij}$ being the covariance between all trajectories $\mathbf{x}_i$ and $\mathbf{x}_j$ (for objects $v_i$ and $v_j$) in the training and validation sets. We determine an ideal threshold $\theta$ so that $A_{ij}=1$ if $R_{ij} > \theta$ and $A_{ij}=0$ otherwise, based on predictive accuracy on the combined training and validation set. $A_{ij}$ denotes the presence of an interaction edge (arbitrary type) between object $v_i$ and $v_j$. We repeat the same procedure for the absolute value of $R_{ij}$, i.e.~$A_{ij}=1$ if $|R_{ij}| > \theta'$ and $A_{ij}=0$ otherwise. Lastly, we pick whichever of the two ($\theta$ or $\theta'$) produced the best match with the ground truth graph (i.e.~highest accuracy score) and report test set accuracy with this setting.
\item \textbf{Corr.~(LSTM)}: Here, we train a two-layer LSTM with shared parameters and 256 hidden units that models each trajectory individually. It is trained to predict the position and velocity for every time step directly and is conditioned on the previous time steps. The input to the model is passed through a two-layer MLP (256 hidden units and ReLU activations) before it is passed to the LSTM, similarly we pass the LSTM output (last time step) through a two-layer MLP (256 hidden units and ReLU activation on the hidden layer). We provide ground truth trajectory information as input at every time step. We train to minimize MSE between model prediction and ground truth path. We train this model for 10 epochs and finally apply the same correlation matrix procedure as in Corr.~(path), but this time calculating correlations between the output of the second LSTM layer at the last time step (instead of using the raw trajectory features). The LSTM is only trained on the training set. The optimal correlation threshold is estimated using the combined training and validation set.
\item \textbf{NRI (sim.)}: In this setting, we replace the decoder of the NRI model with the ground-truth simulator (i.e.~the integrator of the Newtonian equations of motion). We implement both the charged particle and the springs simulator in PyTorch which gives us access to gradient information. We train the overall model with the same settings as the original NRI (learned) model by backpropagating directly through the simulator. We find that for the springs simulation, a single leap-frog integration step is sufficient to closely approximate the trajectory of the original simulation, which was generated with 100 leap-frog steps per time step. For the charged particle simulation, 100 leap-frog steps per time step are necessary to match the original trajectory when testing the simulation decoder in isolation. We find, however, that due to the force clipping necessary to stabilize the original charged particle simulation, gradients will often become zero, making model training difficult or infeasible.
\item \textbf{Supervised}: For this baseline, we train the encoder in isolation and provide ground-truth interaction graphs as labels. We train using a cross-entropy error and monitor the validation accuracy (edge prediction) for model checkpointing. We train with dropout of $p=0.5$ on the hidden layer representation of every MLP in the encoder model, in order to avoid overfitting.
\end{itemize}

\paragraph{Path prediction experiments} Here, we use the following baselines along with our NRI (learned) model:
\begin{itemize}
\item \textbf{Static}: This baseline simply copies the previous state vector $\bx^{t+1}=\bx^t$.
\item \textbf{LSTM (single)}: Same as the LSTM model in Corr.~(LSTM), but trained to predict the state vector difference at every time step (as in the NRI model). Instead of providing ground truth input at every time step, we use the same training protocol as for an NRI model with recurrent decoder (see main paper).
\item \textbf{LSTM (joint)}: This baseline differs from LSTM (single) in that it concatenates the input representations from all objects after passing them through the input MLP. This concatenated representation is fed into a single LSTM where the hidden unit number is multiplied by the number of objects---otherwise same setting as LSTM (single). The output of the second LSTM layer at the last time step is then divided into vectors of same size, one for each object, and fed through the output MLP to predict the state difference for each object separately. LSTM (joint) is trained with same training protocol as the LSTM (single) model.
\item \textbf{NRI (full graph)}: For this model, we keep the latent graph fixed (fully-connected on edge type 2; note that edge types are exclusive, i.e.~edges of type 1 are not present in this case) and train the decoder in isolation in the otherwise same setting as the NRI (learned) model.
\item \textbf{NRI (true graph)}: Here, we train the decoder in isolation and provide the ground truth interaction graph as latent graph representation.
\end{itemize}

\subsection{Motion capture data experiments}
Our extracted motion capture dataset has a total size of 8,063 frames for 31 tracked points each. We normalize all features (position/velocity) to maximum absolute value of 1. Training and validation set samples are 49 frames long (non-overlapping segments extracted from the respective trials). Test set samples are 99 frames long. In the main paper, we report results on the last 50 frames of this test set data.

We choose the same hyperparameter settings as in the physical simulation experiments, with the exception that we train models for 200 epochs and with a batch size of 8. Our model here uses an MLP encoder and an RNN decoder (as the dynamics are not Markovian). We further take samples from the discrete distribution during the forward pass in training and calculate gradients via the concrete relaxation.  The baselines are identical to before (path prediction experiments for physical simulations) with the following exception: For LSTM (joint) we choose a smaller hidden layer size of 128 units and train with a batch size of 1, as the model did otherwise not fit in GPU memory.

\subsection{NBA experiments}
For the NBA data each example is a 25 step trajectory of a pick and roll (PnR) instance, subsampled from the original 25 frames-per-second SportVU data. Unlike the physical simulation where the dynamics of the interactions do not change over time and the motion capture data where the dynamics are approximately periodic, the dynamics here change considerably over time. The middle of the trajectory is, more or less, the pick and roll itself and the behavior before and after are quite different. This poses a problem for fair comparison, as it is problematic to evaluate on the next time steps, i.e.~after the PnR event, since they are quite different from our training data. Therefore in test time we feed in the first 17 time-steps to the encoder and then predict the last 8 steps.

If we train the model normally as an autoencoder, i.e. feeding in the first $N=17$ or $25$ time-steps to the encoder and having the decoder predict the same $N$, then this creates a large difference between training and testing setting, resulting in poor predictive performance. This is expected, as a model trained with $N=17$ never sees the post-PnR dynamics and the encoder trained with $N=25$ has a much easier task than one trained on $N=17$. Therefore in order for our training to be consistent with our testing, we feed during training the first 17 steps to the encoder and predict all 25 with the decoder.

We used a CNN encoder and RNN decoder with two edge types to have comparable capacity to the full graph model. If we ``hard code" one edge type to represent ``non-edge" then our model learns the full graph as all players are highly connected. We also experimented with 10 and 20 edge types which did not perform as well on validation data, probably due to over-fitting.

\end{document}